\documentclass[11pt,twocolumn]{article}

\usepackage[margin=1in]{geometry}
\usepackage{times}          
\usepackage{microtype}      
\usepackage{graphicx}
\usepackage{amsmath,amssymb}
\usepackage{booktabs}
\usepackage{hyperref}

\usepackage{natbib}

\usepackage{microtype}
\usepackage{graphicx}
\usepackage{subcaption}
\usepackage{booktabs} 

\usepackage{hyperref}

\usepackage{multirow}

\usepackage{makecell}

\usepackage{amsmath}
\usepackage{amssymb}
\usepackage{mathtools}
\usepackage{amsthm}
\usepackage{arydshln}
\usepackage{wrapfig}

\usepackage{titlesec}

\titlespacing{\section}{0pt}{1.0ex plus 0.3ex minus 0.2ex}{0.6ex}
\titlespacing{\subsection}{0pt}{0.8ex plus 0.2ex minus 0.2ex}{0.4ex}
\titlespacing{\subsubsection}{0pt}{0.6ex plus 0.2ex minus 0.1ex}{0.3ex}
\titlespacing{\paragraph}{0pt}{0.6ex plus 0.2ex minus 0.1ex}{0.3ex}

\usepackage[capitalize,noabbrev]{cleveref}

\theoremstyle{plain}

\theoremstyle{definition}

\theoremstyle{remark}

\usepackage[textsize=tiny]{todonotes}

\title{Team, Then Trim: An Assembly-Line LLM Framework\\for High-Quality Tabular Data Generation}

\author{
  Congjing Zhang\textsuperscript{a}\thanks{Equal contribution.} \quad
  Ryan Feng Lin\textsuperscript{a}\footnotemark[1] \quad
  Ruoxuan Bao\textsuperscript{b} \quad
  Shuai Huang\textsuperscript{a}\thanks{Corresponding author.} \\[8pt]
  \textsuperscript{a}Department of Industrial \& Systems Engineering, University of Washington\\
  \textsuperscript{b}Department of Management, Shanghai University\\[6pt]
  \texttt{\{congjing, ryanflin\}@uw.edu} \quad \texttt{chris20030909@gmail.com}\\ \texttt{shuaih@uw.edu}
}

\date{} 

\begin{document}
\maketitle





\begin{abstract}
While tabular data is fundamental to many real-world machine learning (ML) applications, acquiring high-quality tabular data is usually labor-intensive and expensive. Limited by the scarcity of observations, tabular datasets often exhibit critical deficiencies, such as class imbalance, selection bias, and low fidelity. To address these challenges, building on recent advances in Large Language Models (LLMs), this paper introduces Team-then-Trim (T$^2$), a framework that synthesizes high-quality tabular data through a collaborative team of LLMs, followed by a rigorous three-stage plug-in data quality control (QC) pipeline. In T$^2$, tabular data generation is conceptualized as a manufacturing process: specialized LLMs, guided by domain knowledge, are tasked with generating different data components sequentially, and the resulting products, i.e., the synthetic data, are systematically evaluated across multiple dimensions of QC. Empirical results on both simulated and real-world datasets demonstrate that T$^2$ outperforms state-of-the-art methods in producing high-quality tabular data, highlighting its potential to support downstream models when direct data collection is practically infeasible.
\end{abstract}

\section{Introduction} 
Machine learning (ML) systems in domains like healthcare \citep{provost2022health}, transportation \citep{washington2020statistical}, and the social sciences \citep{aneshensel2012theory} depend heavily on tabular datasets. According to uniform convergence theorems and generalization bounds in classical statistical learning theory \citep{feldman2018generalization}, if we can sample a sufficiently large number of tabular data points from the underlying space, we can reliably train predictive models that generalize well and achieve accurate performance. In practice, however, we often have to begin with a tabular dataset sampled from the true data distribution, which is typically small and may suffer from various deficiencies, such as bias \citep{little2019statistical}, imbalance \citep{thabtah2020data}, and noise \citep{gupta2019dealing}. For example, in medical studies, such as those on type 1 diabetes, patients may constitute only a small minority compared to non-patient participants, while all of them are at some level of risk, as indicated by the screening results. Under such data imbalance conditions, trained downstream models often struggle to capture the true decision boundary, leading to biased predictions, poor generalization to minority groups, and reduced robustness \citep{chen2023ai}. If we can generate high-quality data that are plausible under domain constraints and diverse enough to cover rare or unseen configurations, the accuracy and robustness of the downstream models can be improved without costly data collection.



However, traditional data augmentation and generation methods, such as resampling (e.g., SMOTE \citep{chawla2002smote}), can only extrapolate from observed samples. While effective within the boundaries of the original data, they remain constrained in their exploration, often reinforcing the same biases and failing to recover rare or missing subpopulations \citep{li2019repair}. Deep generative models, e.g., CTGAN and TVAE \citep{xu2019modeling}, although more expressive, typically demand substantial amounts of data and struggle in low-data regimes where augmentation is most needed. More recently, Large Language Models (LLMs) have emerged as a new paradigm for data generation, leveraging broad world knowledge and few-shot reasoning abilities \citep{patel2024datadreamer,lin2025crowdllm}. CLLM \citep{seedat2024curated} treats the LLM as an unstructured generator: a single LLM receives a flat prompt, produces full samples that are independently and identically distributed, and relies on learning dynamics for curation. However, a single LLM used in isolation is prone to inconsistencies \citep{zhang2025alarm,Zhao_2025_CVPR}. It ignores structural dependencies among features, violates logical constraints, and cannot holistically capture the full complexity of tabular datasets. Tabby \citep{cromp2026tabby} employs a Mixture-of-Experts language model head for each column, but still treats columns independently and relies on architectural modifications and LLM fine-tuning. These requirements limit its flexibility, increase deployment costs, and reduce practicality in low-data settings. Further related work is provided in Appendix \ref{appendix_sec_relatedwork}. The issues in current work limit their reliability, especially when tasked with generating multi-faceted records from small datasets.

Beyond generation, ensuring the quality of synthetic data is equally critical. LLMs are known to hallucinate \citep{bang2025hallulens}, introduce implausible entries \citep{yao2023llm}, or produce distributions that deviate from domain requirements \citep{xu2024hallucination}. Existing methods typically focus on plausibility checks or heuristic filtering \citep{sousa2024generation}, but they often overlook deeper dimensions of quality, such as consistency with downstream objectives, fidelity to minority subgroups, and diversity of coverage. Without a principled quality control (QC) process, synthetic data risks amplifying noise rather than mitigating it.

To address these two challenges, we introduce \textit{Team-then-Trim} (T$^2$), a framework that combines teaming for structured data generation with trimming for rigorous three-stage QC. Unlike monolithic generators, our framework decomposes a dataset into semantically coherent components and assigns each to a specialized LLM worker with a clearly defined role. Workers operate sequentially, conditioning on previously produced components to preserve inter-feature logic and domain constraints. T$^2$ produces data in batches, allowing large volumes of synthetic data to be created with lower costs and higher efficiency \citep{citovsky2021batch, ren2021survey}, while subjecting each batch to a rigorous QC process. The QC pipeline consists of three stages: (1) a sanity check validates variable ranges, categorical consistency, and inter-feature dependencies; (2) objective-related cost assessment conducts a model-based duel between LLM-generated batches and bootstrap samples from the original data, keeping low-residual points, and (3) diversity inspection with clustering-based coverage checks admit batches expanding coverage without skew. Together, these stages transform an initial synthetic pool into a task-aligned, diverse, and constraint-consistent dataset for downstream learning. The QC pipeline is modular and can be plugged into any generative approach. 

To sum up, the contributions of this paper are threefold: (1) We introduce T$^2$, a novel framework that conceptualizes high-quality tabular data generation as a product manufacturing process, where raw generated data is synthesized in batches through the collaborative assembly of specialized LLM generators (teaming) and subsequently refined through a QC process (trimming). (2) We propose a three-stage plug-in QC pipeline that transforms raw synthetic batches into high-quality data using only a small original dataset, improving validity, learnability, informativeness, and diversity. (3) We conduct extensive experiments on simulated and real-world data to demonstrate that T$^2$ can reliably generate high-quality tabular data across diverse metrics and solve data scarcity, imbalance, incompleteness, and noise.

\begin{figure*}[!t]
    \centering
        \includegraphics[width=0.85\textwidth]{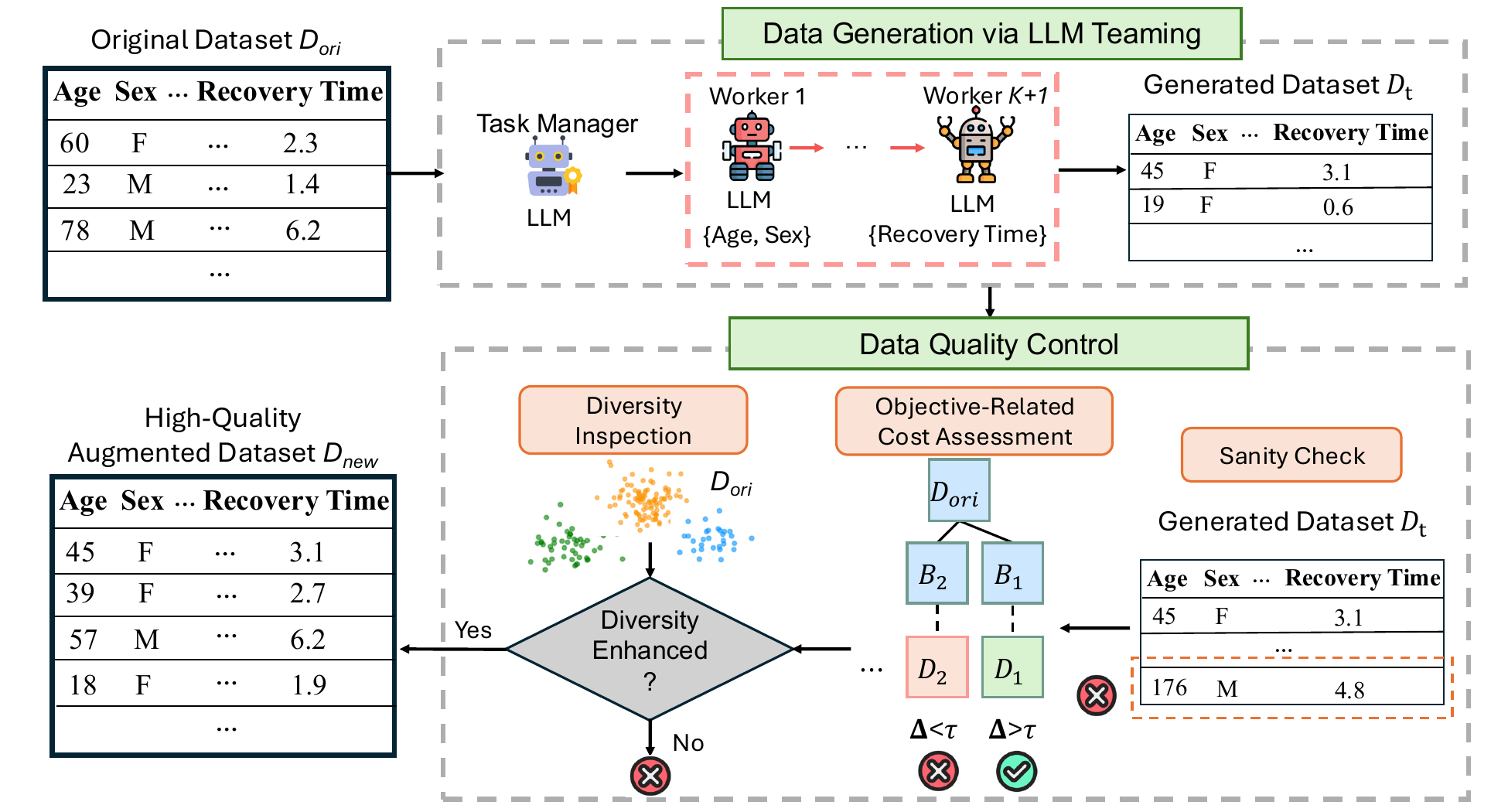}
\caption{An illustration of our T$^2$ framework. LLM teaming is for structured data generation and trimming is for rigorous three-stage QC. }\label{framework}
\end{figure*}

\section{Team-then-Trim Framework}
We now describe our T$^2$ framework for generating high-quality tabular data that can be used by a broad spectrum of downstream predictive models. Consider the tabular data space $\mathcal{X}\times\mathcal{Y}$ equipped with a probability measure induced by the true data distribution $p_\text{true}$. Here, $\mathcal{X}=\mathcal{X}_1\times\cdots\times\mathcal{X}_d$ denotes the $d$-dimensional feature space where $\mathcal{X}_j$ is the $j$-th feature dimension, and $\mathcal{Y}$ denotes the label space. Our framework starts with a small tabular dataset sampled from $p_\text{true}$, denoted by $D_\text{ori}=\{(\boldsymbol{x}_{i},y_i)\}_{i=1}^{n}$, where $\boldsymbol{x}_{i}=(x_{i,1},\cdots,x_{i,d})\in\mathcal{X}$ and $y_{i}\in\mathcal{Y}$. It generates $T$ additional batches of new data $D_t,t=1,\cdots,T$ without explicit knowledge of $p_\text{true}$. The QC pipeline examines the generated data. Let $\mathcal{T}_{\text{kept}} \subseteq \{1,\ldots,T\}$ denote the index set of generated batches that pass the QC pipeline and $D_{\text{kept}} = \bigcup_{t \in \mathcal{T}_{\text{kept}}} D_t$. Our goal is to construct an augmented dataset $D_{\text{new}} = D_{\text{ori}} \cup D_{\text{kept}}$ so that a downstream predictive model $y=f(\boldsymbol{x})$ trained on $D_\text{new}$ achieves improved generalization performance compared to training exclusively on $D_\text{ori}$. An overview of the T$^2$ framework is shown in Figure \ref{framework}. We present details in the following subsections.

\subsection{Raw Data Generation via LLM Teaming}\label{sec2.1}
An assembly line in manufacturing typically breaks down a complex production process into a sequence of smaller subtasks executed by specialized workers under the coordination of a task manager. Analogously, the T$^2$ framework employs a team of pretrained LLMs, where a task manager and multiple workers collaborate to produce a raw tabular dataset. Given a data generation task, the task manager LLM, guided by its domain knowledge of the specific task and its analysis of the real data $D_\text{ori}$, first decomposes the data structure into multiple components. It then assigns component-generation subtasks to different worker LLMs and orchestrates their execution according to a specified topology. The workers perform their assigned tasks to generate the components sequentially, following the instructions given by the task manager LLM in an assembly-line fashion. 

\begin{figure}[!t]
  \centering
  \includegraphics[width=0.8\columnwidth]{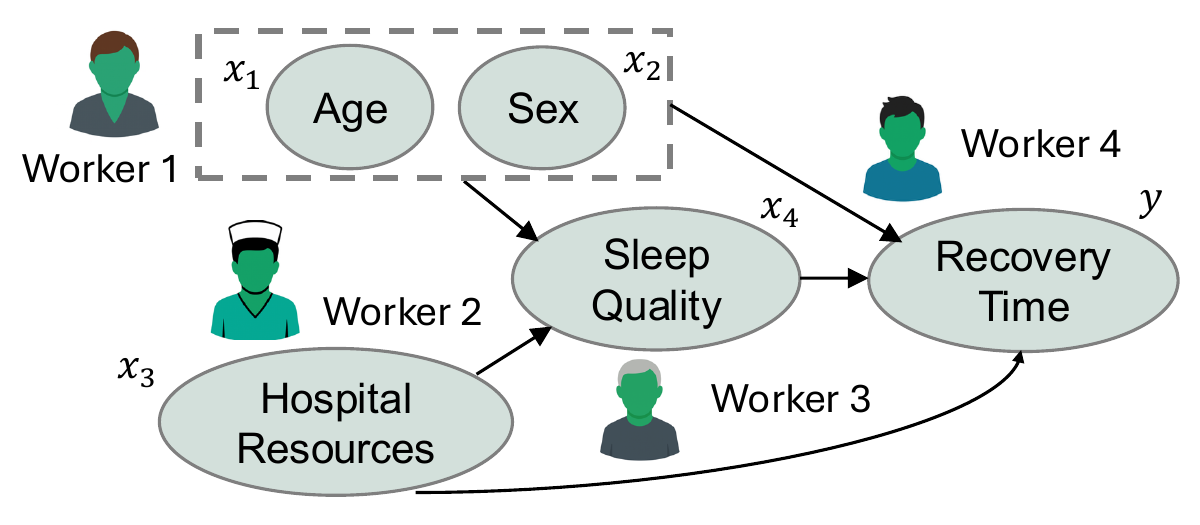}
  \caption{An example of task coordination.}
  \label{fig2}
\end{figure}

\textbf{Task manager.} Given the original dataset $D_\text{ori}$, the task manager LLM $\Phi$ is additionally prompted with the feature dictionary to facilitate understanding of the task (see Appendix \ref{app_prompt_exm}). Based on its understanding and domain knowledge, it will partition the full feature space $\mathcal{X}$ into $K$ disjoint components, i.e., $\mathcal{X}=\mathcal{X}_{I_1}\times\cdots\times\mathcal{X}_{I_K}$, where $\mathcal{X}_I=\prod_{i\in I}\mathcal{X}_i$ is the Cartesian product of all the single feature spaces whose indices belong to $I$. Thus, the partition should satisfy $\bigcup_{k=1}^{K} I_k = \{1, \dots, d\}$ and for $\forall k,l = 1,\dots,K, k \neq l$, $I_k \cap I_l = \emptyset$. This partition divides the data generation workload into $K$ components, allowing different worker LLMs to handle distinct parts of the task. Each component represents a coherent semantic category, with all the features sharing underlying semantic characteristics though not necessarily correlated. For example, consider the task of predicting post-surgery recovery time with four predictors: age, sex, hospital resources, and sleep quality. As Figure \ref{fig2} shows, these predictors can be grouped into three categories: demographic features (age and sex), an independent environmental feature (hospital resources), and a feature potentially influenced by both demographic and environmental factors (sleep quality). Without incorporating such structural knowledge, the generated data may overlook complex dependencies among features and fail to capture their semantically meaningful relationships. In our T$^2$ framework, the task manager LLM fully utilizes such knowledge to coordinate data generation. It views the different components as nodes in a graph, resulting in an activity-on-vertex network (precedence graph \citep{boffey1982project}) $G=(V, E)$ explicitly encoding their relationships, where $V=\{I_1,\cdots, I_K\}$ and $E$ encodes the information of the graph links. Thus, the work of the task manager LLM can be expressed as: 
\begin{align}
G=\Phi(\mathcal{P}(D_\text{ori}, \boldsymbol{\xi})),
\end{align}
where $\boldsymbol{\xi}$ is contextual information of the dataset (e.g., variable dictionary) and $\mathcal{P}$ denotes the encoding function that turns the input information into prompts. This scheduling enables parallel generation of components without interdependencies and sequential generation when dependencies exist, reducing per-component complexity and improving semantic coherence. With less information required to generate components, the overall process is able to produce more precise and semantically coherent data.



\paragraph{Specialized workers.} Our T$^2$ framework employs a set of LLMs $\Psi_1,\cdots,\Psi_K$ as specialized workers to efficiently and consistently generate the $K$ different data components defined by the task manager LLM. All the specialized workers follow the coordination instructions provided by the task manager to generate data sequentially in the prescribed order. Each worker receives a prompt containing specific information about the component it is responsible for, along with all previously generated data components on which its assigned component may depend. The generated data component can be expressed, for $\forall k=1,\cdots,K$, as:
\begin{align}
\mathbf{X}_k\sim\Psi_k(\mathcal{P}_k(D_\text{ori}, \boldsymbol{\xi}_k,G)|\bigcup_{l\in\text{pa}(I_k,G)}\mathbf{X}_l),
\end{align}
where $\mathcal{P}_k$ is the individual prompt encoding function, $\text{pa}(I_k, G)$ denotes the set of parent nodes of $I_k$ on the graph $G$ and $\boldsymbol{\xi}_k$ is the contextual information about the features in the $k$-th component. Thus, the parent components, the generation schedule determined by the task manager, and the original data with contextual information together constitute the foundation for generating $\mathbf{X}_k$. For example, in Figure \ref{fig2}, to generate the data column for the feature sleep quality, workers 1 and 2 should generate the demographic feature component (including age and sex) and the environmental feature component (i.e., hospital resources) beforehand, as this guarantees that the subsequent generation of sleep quality does not contradict the domain knowledge encoded within the worker LLM. It is important to note that, due to the stochastic nature of LLMs, the generation process is inherently nondeterministic. Consequently, the generated data is not fixed even with identical prompts or conditions. Once the parent components have been generated by upstream worker LLMs under the coordination of the task manager, the specialized worker proceeds to sample candidate data points from an underlying, unknown distribution as instructed by the received prompt. Since the features in the same component semantically belong to the same category, each worker can focus on generating semantically related concepts, which makes the generation process more efficient and reliable. Taking advantage of the prior knowledge, contextual understanding, and in-context learning capabilities of LLMs, the generated data is expected to exhibit higher internal consistency and factual fidelity \citep{saglam2025large, kozlowski2025semantic, wang2025semantic} through the teaming of these workers, which is a notable advantage over a single LLM that handles multiple semantic categories simultaneously. An example that illustrates such an advantage is shown in Appendix \ref{expteam}. Finally, when all these workers finish generating data, we aggregate all the generated components and assemble them into a full design (unlabeled dataset) through a concatenation function:
\begin{align}
\mathbf{X}_\text{gen}=\text{Concat}(\mathbf{X}_1,\cdots,\mathbf{X}_K).
\end{align}
An additional worker LLM $\Psi_{K+1}$ takes as input the unlabeled dataset and generates labels for the design:
\begin{align}
    \boldsymbol{y}_\text{gen}\sim \Psi_{K+1}(\mathcal{P}_y(D_\text{ori},\boldsymbol{\xi},\mathbf{X})),
\end{align}
where $\mathcal{P}_y$ denotes the prompt. Here, the labels can be restricted to meaningful values by encoding the constraints in the context $\boldsymbol{\xi}$. For example, for binary-labeled data, the label worker is prompted to produce values from $\{0,1\}$. If this restriction is not required, the worker is allowed to generate different labels beyond the ones in the given original dataset to encourage its exploration. With the labels generated and all components combined, we obtain a complete dataset $\mathcal{D}_t=\{\mathbf{X}_{\text{gen},t},\boldsymbol{y}_{\text{gen},t}\}$ as the $t$-th generated batch.

\subsection{Data Quality Control} \label{sec3}




Even though the generated data may appear plausible, it can still suffer from quality issues due to LLM hallucination or mistakes \citep{huang2025survey, chen2024deep, liu2024exploring}. Just as physical products require QC to ensure their reliability, the raw products of the data generation assembly line, namely, $D_t$, must also undergo a quality assurance process to ensure they are readily usable by various downstream applications. Owing to their black-box nature, it is difficult to manipulate the generation process of worker LLMs or exert direct control over data quality during generation, unless we perform full training or fine-tuning with large amounts of ground-truth data, which typically incurs substantial costs. Consequently, quality assurance efforts are preferably concentrated in the post-generation stage, which motivates the design of our three-stage plug-in QC pipeline.

In our pipeline, QC is applied batch-wise, trimming or discarding data as necessary. Compared with single-point inspection, batch-level QC has multiple advantages, such as reducing cost and improving efficiency \citep{citovsky2021batch, ren2021survey}. Moreover, batches provide richer collective information than isolated samples, which enables more effective and informative quality inspection \citep{kirsch2019batchbald}. Assume the batch size is $n_b$. If a batch meets the predefined quality criteria, QC trims the unsatisfactory samples in the batch and integrates the remaining data into the existing dataset $D$ (initially $D_\text{ori}$) to construct a new set; otherwise, the entire batch is discarded. This iterative generation-and-evaluation cycle continues until sufficient data are admitted. For evaluation, batch quality is assessed from three complementary aspects: 1) \textit{Sanity check}. The batch is examined with respect to its variables to filter out clearly invalid or impossible data samples. 2) \textit{Objective-related cost assessment}. Low-quality data are identified and rejected based on the batch's total cost related to the learning objective of a given model. 3) \textit{Diversity inspection}. The batch is further checked to ensure sufficient coverage of the data space. 

\paragraph{Sanity check.} To ensure the samples in the batch $D_t$ are valid, we perform a sanity check based on the types and values of features together with necessary relational constraints. Let the full feature index set be $S=\{1,2,\cdots, d\}$. For continuous features indexed by $S_1=\{k_1,\cdots, k_m\}\subset S$ where $k_i\neq k_j,\forall i,j$, their values in all samples of $D_t$ should be within a reasonable range. This constraint can be characterized by a disjunction of linear inequalities $\Omega=\Omega_1\lor \cdots\lor\Omega_m$, where $\Omega_i$ denotes a boundary inequality over the feature in the form $l_i\leq x\leq u_i$. Here, $l_i$ and $u_i$ denote the lower and upper bounds of the admissible range for the $i$-th continuous feature, respectively. For categorical features indexed by $S_2=S\backslash S_1$, we ensure their values fall within the allowable categories $C_1,\cdots,C_{n_{l}}$ for any feature indexed by $l\in S_2$. By introducing dummy variables $x_{\cdot, l,1},\cdots, x_{\cdot, l,n_l}\in\{0,1\}$ where $x_{\cdot, l,j}=\mathbb{I}[x_{\cdot, l}\in C_j], \forall j=1,\cdots, n_l$ denotes the indicator of whether feature $x_{\cdot, l}$ belongs to a category $C_j$, we impose the categorical constraint $\sum_{j=1}^{n_l}x_{\cdot,l,j}=1$. Furthermore, strict relationships among features can be expressed as a similar disjunction of inequalities of the form $\sum_{l\in S_1}w_{l}x_{\cdot,l}+\sum_{l\in S_2}\sum_{j=1}^{n_l}w_{l,j}x_{\cdot,l,j}+b\geq 0$ with coefficients $w_l$ and $w_{l,j}$. Therefore, all requirements on features and their interdependence can be formulated as a unified set of logical and linear constraints. The sanity check then reduces to a constraint satisfaction problem, and any sample violating the predefined constraints is discarded. In practice, owing to the prior knowledge of LLMs, generated batches typically satisfy these constraints when features are easily interpretable. Nevertheless, this check remains indispensable for preventing trivial errors and eliminating semantically meaningless data that compromise downstream tasks.

\paragraph{Objective-related cost assessment.} Even if a data batch passes a sanity check, its utility for downstream tasks is not guaranteed, as it may still suffer from issues such as bias, imbalance, and noise. To filter out such deficient data, additional screening is required to ensure that the generated data can be reliably used for training a predictive model. Thus, we focus on model-based prediction and evaluate the potential cost related to the learning objective compared to the original data. We assume access to the downstream model $y=f_{\theta}(\boldsymbol{x})$ parametrized by $\theta$. To facilitate the assessment, our key idea is to compare the effectiveness of the generated data with the original data in downstream tasks. Specifically, we perform bootstrapping from the original set $D_\text{ori}$ and obtain a batch $B_t$ of the same size as the generated batch $D_t$, i.e., $n_b$. We combine the two batches to build a mixture $\tilde{D}_t=D_t\cup B_t$ and make predictions on all samples in $\tilde{D}_t$ with $f_{\theta}$. Thus, for $\boldsymbol{x}_i$ in $\tilde{D}_t$ with label $y_i$, we have the per-sample cost for classification: 
\begin{align}
    r_i=1-p_f(y_i|\boldsymbol{x}_i),
\label{residual}
\end{align}
where $p_f$ is the predicted probability score given by the model $f_{\theta}$. On the one hand, to build a high-quality batch of size $n_b$, we aim to minimize the prediction cost. On the other hand, excessive cost minimization of the model on a small dataset may cause overfitting and potentially degrade the model’s generalization performance. Instead of pursuing either extreme, we select $2(1-\alpha)n_b$ samples in $\tilde{D}_t$ to build an updated batch. Here, $\alpha\in(0,1)$ controls the sample size, which is set to 0.5 in practice for a fair comparison with $B_t$. We sort the costs in Eq. (\ref{residual}) for all the samples in $\tilde{D}_t$ and build an empirical distribution $F$. Denote $F^{-1}(\beta)\triangleq\inf\{e: F(e)\geq \beta\}$ as the $\beta$-quantile of $\{r_i\}_{i=1}^{2n_b}$. For a specific $\beta\in(0,0.5)$, we select the samples indexed by $J=\{i_1,\cdots,i_{n_b}\}$ from $\tilde{D}_t$ which satisfy $r_i\in[F^{-1}(\beta), \,\,F^{-1}(1-\alpha+\beta)],\forall i\in J$. These selected samples form a refined dataset $\overline{D}_t=\{(\boldsymbol{x}_i,y_i)\}_{i\in J}$, which is further compared with the representative batch of original data $B_t$. We test whether augmenting the existing dataset $D$ with the refined set $\tilde{D}_t$ yields a significant improvement over adding a control batch $B_t$ by comparing two sets:  $D^1_\text{gen}=D\cup\overline{D}_t$ and $D^2_\text{gen}=D\cup B_t$. Following \citet{mehta2022information,smith2023prediction}, the information gain (IG) of combining any dataset $D'$ with $D$ is expressed as:
\begin{align}
\text{IG}(D')=\mathbb{H}(\theta|D)-\mathbb{H}(\theta|D, D'),
\end{align}
where $\mathbb{H}(\cdot)$ is the entropy function. Thus, we can calculate the information gain of the two sets as $IG(\overline{D}_t)$ and $IG(B_t)$, and compute their gap as $\Delta=IG(\overline{D}_t)-IG(B_t)$, which measures the marginal information gain from the generated data batch $\overline{D}_t$ over the bootstrapped original data batch $B_t$. We only keep the batch of generated data when it provides sufficient information, i.e., $\Delta$ is greater than a threshold $\tau$. If $\Delta>\tau$, we merge the updated batch $\overline{D}_t$ into the existing dataset $D$, otherwise we discard it.

\paragraph{Diversity inspection.} 
Beyond validity and objective alignment, it is crucial to ensure that the generated data enhances (or at least maintains) the diversity of $D_\text{ori}$, especially for rare subpopulations. We uncover the underlying structure of $D_\text{ori}$ by performing clustering. The optimal number of clusters is determined by maximizing the average silhouette score of all the data points in $D_\text{ori}$. The silhouette score $s(\boldsymbol{x}_i)$ of a data point $\boldsymbol{x}_i$ can be computed as $s(\boldsymbol{x}_i)=\frac{z^\text{coh}(\boldsymbol{x}_i)-z^\text{sep}(\boldsymbol{x}_i)}{\max\{z^\text{coh}(\boldsymbol{x}_i), z^\text{sep}(\boldsymbol{x}_i)\}}$.
Here, $z^\text{coh}(\boldsymbol{x}_i)$ denotes the average distance between $\boldsymbol{x}_i$ and all other data points within the same cluster, and $z^\text{sep}(\boldsymbol{x}_i)$ denotes the smallest average distance between $\boldsymbol{x}_i$ and all points in the nearest distinct cluster, which can be expressed as $z^\text{coh}(\boldsymbol{x}_i)= \frac{1}{S(c_i) - 1} \sum_{j:c_j=c_i, j \neq i} d(\boldsymbol{x}_i, \boldsymbol{x}_j)$ and $z^\text{sep}(\boldsymbol{x}_i)= \min_{c \neq c_i}  \frac{1}{S(c)} \sum_{j:c_j=c} d(\boldsymbol{x}_i, \boldsymbol{x}_j) \,$. Here, $c_i$ refers to the cluster to which $\boldsymbol{x}_i$ belongs, $S(c)$ denotes the number of data points belonging to the cluster $c$, and $d(\boldsymbol{x}_i,\boldsymbol{x}_j)$ is the Euclidean distance between $\boldsymbol{x}_i$ and $\boldsymbol{x}_j$. After partitioning $D_\text{ori}$ into $M$ clusters, we train a multiclass classifier (i.e., an MLP) using the cluster assignments as labels. This classifier effectively learns to map any data point from the feature space to one of the identified data-space regions. If the batch of generated data $D_{t}$ passes the objective-related cost assessment, each generated observation is passed through the MLP to be categorized into one of the baseline clusters. The classifier acts as a diversity proxy, allowing us to project newly generated data into the same structural framework as the original dataset. 
For any dataset $D$, let $p_D(m)= \frac{1}{|D|}\sum_{i\in D}\mathbb{I}[c_i=m]$, $m=1,\cdots,M$, and the cluster entropy $\mathbb{H}_c(D) = -\sum_{m=1}^M p_D(m)\,\log p_D(m)$. We quantify the diversity gain of a generated batch $D_t$ by the entropy increase:
\begin{align}
\Delta \mathbb{H}_c(D_t) = \mathbb{H}_c(D_{\text{ori}}\cup D_t)\;-\;\mathbb{H}_c(D_{\text{ori}}).
\end{align}
A generated batch $D_{t}$ is accepted only if the entropy improvement of the combined dataset ($D_\text{ori} \cup D_t$) over the original dataset ($D_\text{ori}$) exceeds a predefined threshold. Intuitively, entropy increases when synthetic samples expand coverage by populating underrepresented clusters or distributing more evenly across clusters, and decreases when they concentrate on dominant modes. Accordingly, this criterion retains only batches that meaningfully broaden coverage of the feature manifold without introducing substantial distributional skew. Our notion of diversity emphasizes coverage of the feature space rather than the mere class balance. Computing entropy over outcome labels primarily reflects whether the generator produces a balanced class distribution, which can appear ``diverse" even when diversity is superficial, e.g., when labels are flipped without introducing new or meaningful feature patterns. In contrast, cluster-based entropy explicitly rewards the dispersion of samples across distinct regions of the feature manifold, capturing structural diversity that is more relevant for downstream learning. 






\section{Experiments and Results}\label{exp_sec}
We conduct a comprehensive evaluation across simulated studies and real-world data. In the simulated settings, we recreate common deficiencies in tabular datasets, i.e., imbalance, incompleteness, and noise. For each setting, we also examine performance under data-scarce regimes. For real-world datasets, we examine the downstream utility and data quality. Our code will be released publicly upon acceptance.


\subsection{Setup}
For downstream evaluation, we consider four models: Logistic Regression, SVM, MLP, and Random Forest. We compare T$^2$ against established baselines: EPIC \citep{kim2024epic}, CLLM \citep{seedat2024curated}, TVAE \citep{xu2019modeling}, CTGAN \citep{xu2019modeling}, and SMOTE \citep{chawla2002smote}. Our framework is designed for scenarios where the original data is highly deficient, e.g., too small and incomplete to support reliable training or fine-tuning of large models. Thus, we focus on baselines that operate with frozen LLMs or non-parametric generators. We employ Llama 3.3 70B Instruct as the main backbone LLM for T$^2$, EPIC, and CLLM. We also run experiments using Grok-4.1-Fast and GPT-5.1 in the data imbalance and incompleteness settings for comparison. In the simulated settings, we construct a diabetes prediction dataset \citep{rauba2024self}, and a travel behavior dataset \citep{zhu2020personalized}. For real-world experiments, we adopt the Drug dataset from the UCI repository \citep{fehrman2017five} and COMPAS from OpenML \citep{angwin2016machine}. Except for experiments using $D_\text{ori}$ alone, all results use $D_\text{ori}$ augmented with generated samples as training data and are averaged over 10 random seeds (10 splits) and 4 downstream models. More experimental details are in Appendix \ref{exp_detail}.





\subsection{Simulated Studies}

\paragraph{Data imbalance.}

In the diabetes dataset, we simulate three levels of risk groups, i.e., low risk (LR), moderate risk (MR), and high risk (HR), with a 7:2:1 ratio to create label imbalance. The size of $D_\text{ori}$ is 10, following this imbalanced ratio. Table \ref{AUC_combined} reports the average AUC of different methods. Our T$^2$ framework achieves the strongest and most consistent gains across all groups, which indicates that LLM teaming with QC is particularly effective in addressing label imbalance in tabular data. Evaluation of computational costs is shown in Appendix Table \ref{comp_cost}.

\begin{table}[htbp]
\caption{Average AUC (\%) under imbalance and incompleteness.}
\label{AUC_combined}
\centering
\footnotesize
\setlength{\tabcolsep}{5pt}
\renewcommand{\arraystretch}{0.8}

\begin{tabular}{@{}ccccccc@{}}
\toprule
\multirow{2}{*}{\textbf{Method}} 
& \multicolumn{3}{c}{\textbf{Data Imbalance}} 
& \multicolumn{3}{c}{\textbf{Data Incompleteness}} \\
\cmidrule(lr){2-4} \cmidrule(lr){5-7}
& \textbf{LR} & \textbf{MR} & \textbf{HR}
& \textbf{LR} & \textbf{MR} & \textbf{HR} \\
\midrule
$D_\text{ori}$ & 63.92 & 59.16 & 64.24 & 58.59 & 48.69 & -- \\
SMOTE     & 68.99 & 58.73 & 68.66 & 64.70 & 48.52 & -- \\
TVAE      & 68.79 & 60.75 & 65.22 & 62.81 & 49.85 & -- \\
CTGAN     & 67.30 & 59.61 & 67.00 & 65.14 & 49.87 & -- \\
CLLM      & 64.43 & 57.15 & 65.79 & 63.68 & \textbf{54.47} & 50.49 \\
EPIC      & 62.24 & 51.79 & 75.33 & 64.29 & 50.28 & -- \\
T$^2$ (\textbf{ours}) 
& \textbf{71.55} & \textbf{62.03} & \textbf{76.98}
& \textbf{71.16} & 52.26 & \textbf{64.67} \\
\bottomrule
\end{tabular}
\end{table}

\begin{figure*}[t]
    \centering
    \begin{minipage}{0.7\textwidth}
        \centering
        \begin{subfigure}{0.33\linewidth}
            \centering
            \includegraphics[width=\linewidth]{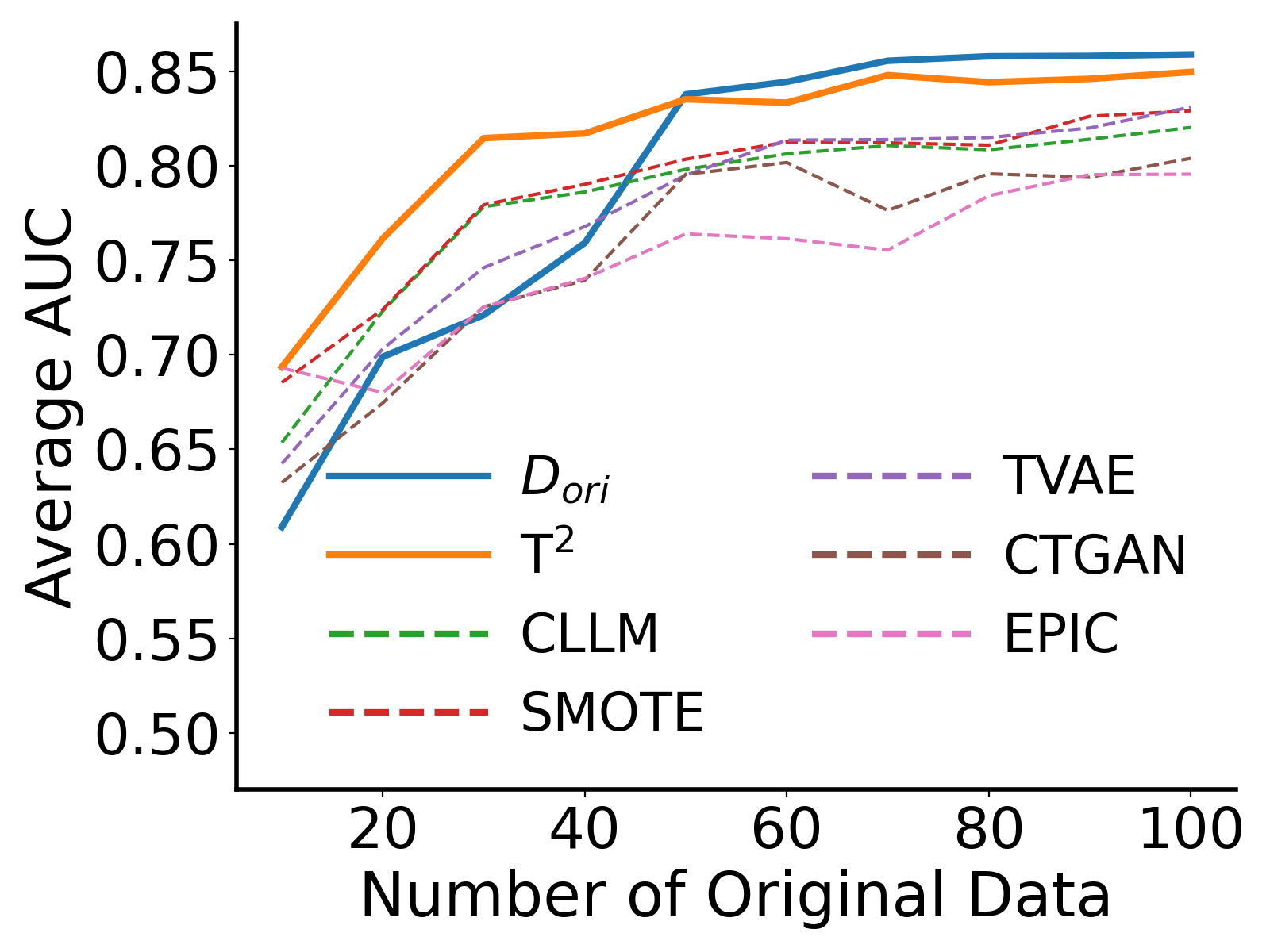}
            \caption{Flip ratio is 0.2.}
            \label{noise_diabetes_0.2}
        \end{subfigure}%
        \begin{subfigure}{0.33\linewidth}
            \centering
            \includegraphics[width=\linewidth]{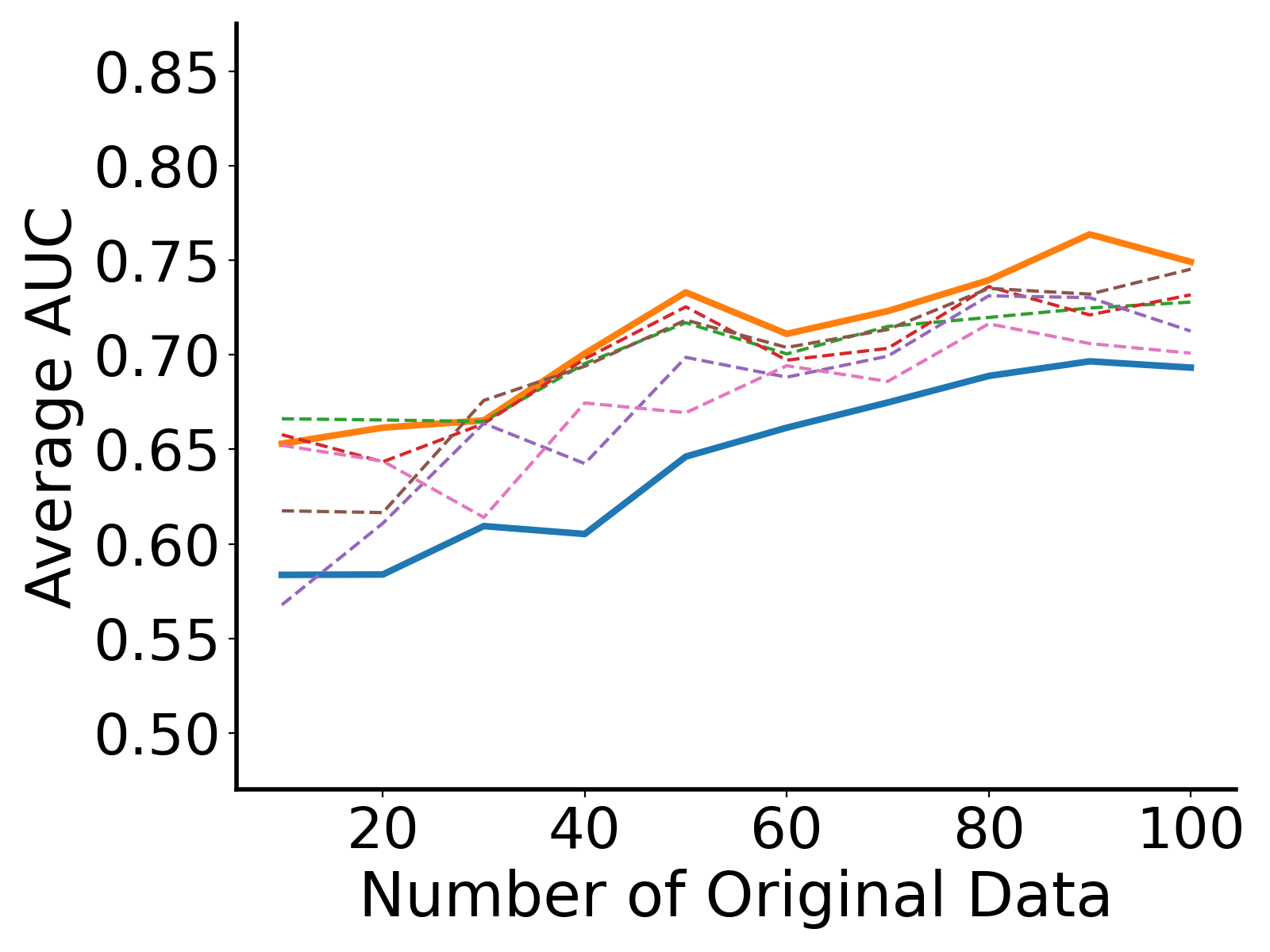}
            \caption{Flip ratio is 0.3.}
            \label{noise_diabetes_0.3}
        \end{subfigure}%
        \begin{subfigure}{0.33\linewidth}
            \centering
            \includegraphics[width=\linewidth]{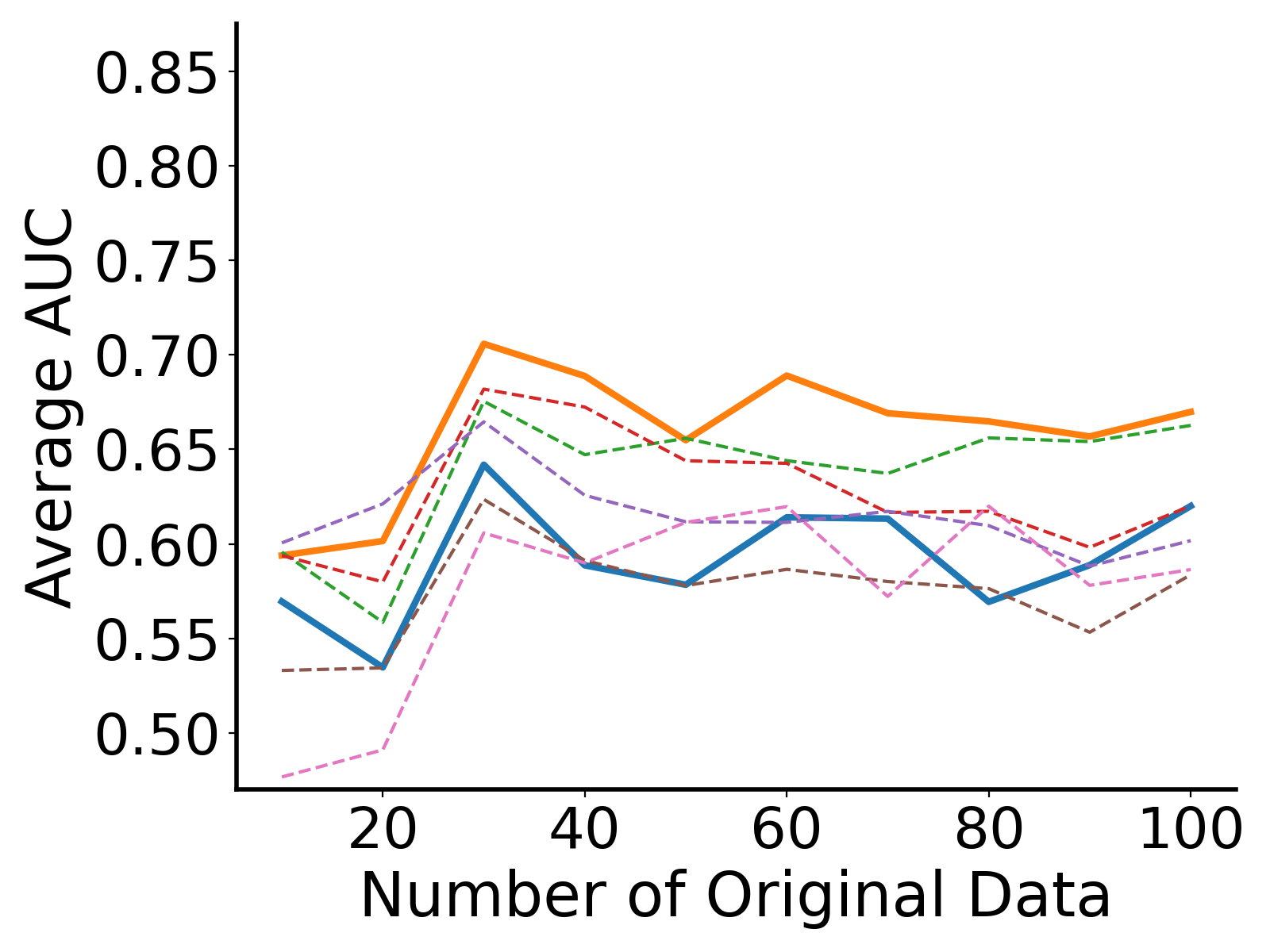}
            \caption{Flip ratio is 0.4.}
            \label{noise_diabetes_0.4}
        \end{subfigure}%
        \caption{Average AUC vs. $|D_\text{ori}|$ under flip ratios of 0.2, 0.3, and 0.4 on the Diabetes dataset.}
        \label{noise_diabetes}
    \end{minipage}\hfill
    \begin{minipage}{0.3\textwidth}
        \centering
        \captionof{table}{Average AUC (\%) under flip ratios of 0, 0.3, and 0.4 on TravelBehavior.}
        \footnotesize
        \setlength{\tabcolsep}{4pt}
        \renewcommand{\arraystretch}{0.8}
        \begin{tabular}{@{}cccc@{}}
        \toprule
        \textbf{Method} & \textbf{No Flip} & \textbf{0.3} & \textbf{0.4} \\
        \midrule
        $D_\text{ori}$ & \textbf{98.55} & 60.61 & 50.02 \\
        SMOTE     & 98.54 & 58.95 & 51.03 \\
        TVAE      & 96.62 & 57.80 & 49.11 \\
        CTGAN     & 91.40 & 58.91 & 49.26 \\
        CLLM      & 97.73 & 62.05 & 53.14 \\
        EPIC      & 98.12 & 51.63 & 50.41 \\
        T$^2$ (\textbf{ours}) & 98.24 & \textbf{63.44} & \textbf{53.38} \\
        \bottomrule
        \end{tabular}
        \label{noise_travel_table}
    \end{minipage}
\end{figure*}

\paragraph{Data incompleteness.}
To simulate real-world incompleteness, we construct $D_\text{ori}$ with LR:MR:HR = 8:2:0, such that the HR group constitutes a missing subpopulation in the training data. The size of $D_\text{ori}$ is set to 10. The test set retains the distribution with LR:MR:HR = 7:2:1. Table \ref{AUC_combined} shows that classical generation baselines completely fail to recover the missing HR subgroup. Although EPIC is an LLM-based method, it struggles to generate samples for unseen classes, as its generation process is tightly conditioned on class-specific examples provided in the prompt and lacks detailed task instructions to support extrapolation to unseen labels. As another LLM-based method, CLLM attains a relatively high AUC in the MR group but generalizes poorly to the HR group, indicating that, without explicit LLM teaming and rigorous QC, the generated samples are prone to noise and misalignment. By contrast, T$^2$ demonstrates a strong performance, with a 14.18\% gain over CLLM in the HR group. t-SNE visualizations are in Appendix \ref{appendix_result}.

\paragraph{Data noise.}
To simulate data noise, we encode the outcome variables in both datasets as binary indicators. Larger flip ratios of the true labels correspond to higher levels of label corruption, i.e., more noise. In the diabetes dataset, at a low noise level (flip ratio = 0.2, shown in Figure \ref{noise_diabetes_0.2}), T$^2$ outperforms all models when the number of original samples is small. As the number of original samples increases, the clean data alone already yields a high AUC due to the mild noise. As noise increases to 0.3 (shown in Figure \ref{noise_diabetes_0.3}) and 0.4 (shown in Figure \ref{noise_diabetes_0.4}), T$^2$ sustains stable gains. This demonstrates the robustness of T$^2$ to label corruption even under moderate to severe noise. Table \ref{noise_travel_table} shows the average AUC under no flip and flip ratios of 0.3, and 0.4 on the TravelBehavior dataset, with 60 original samples. The observed trends mirror those in the diabetes dataset. Overall, T$^2$ enhances downstream utility across noisy settings in both datasets, with particularly strong gains under higher noise levels, highlighting its effectiveness as a generative framework for challenging tabular data.

\subsection{Real-World Data}
\paragraph{Downstream utility.}

\begin{figure}[htbp]
    \centering
    \begin{subfigure}{0.5\columnwidth}
        \centering
        \includegraphics[width=\linewidth]{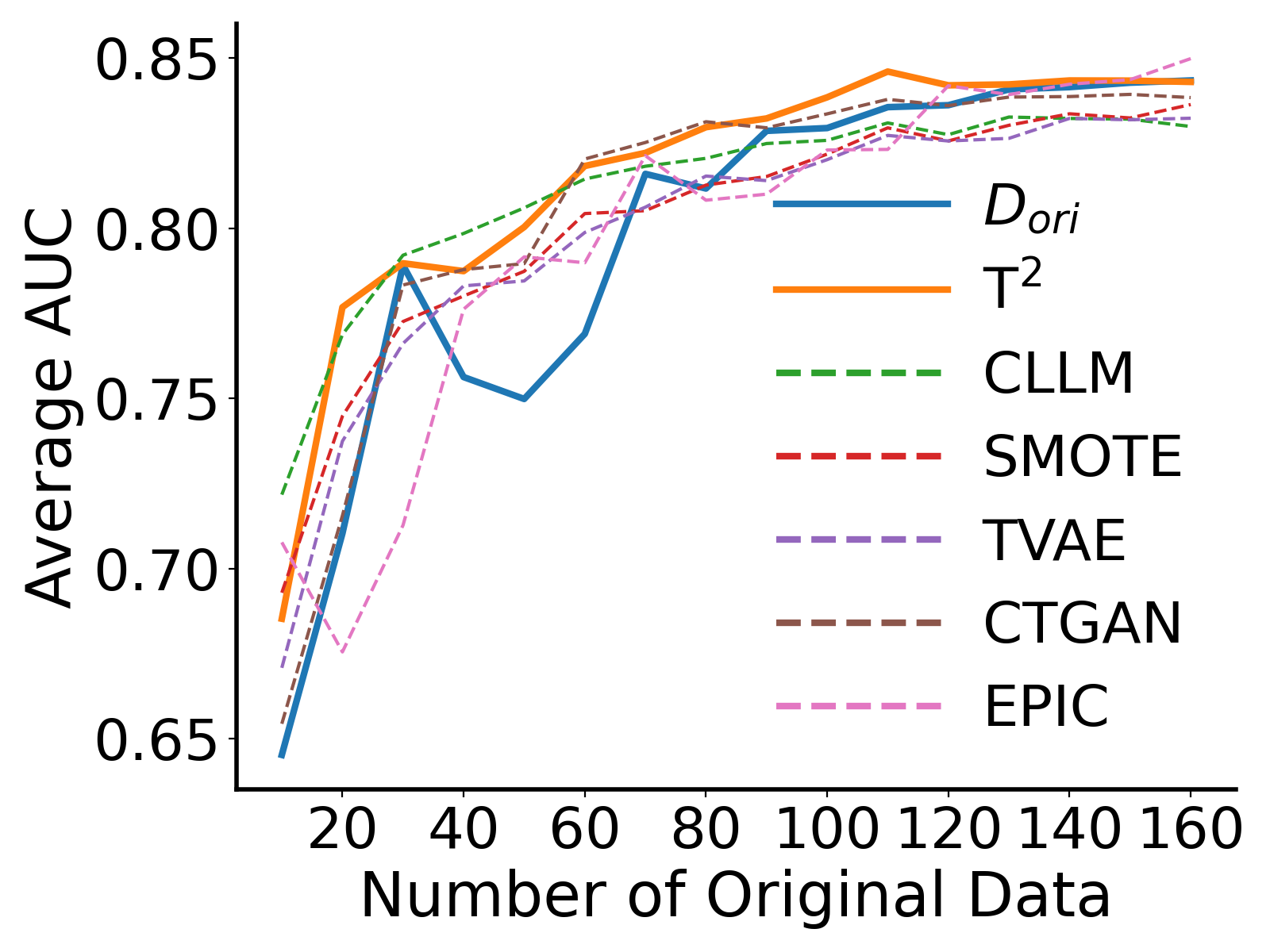}
        \caption{White female, 25-34.}
        \label{drug_1-0-6_whitefemale}
    \end{subfigure}\hfill
    \begin{subfigure}{0.5\columnwidth}
        \centering
        \includegraphics[width=\linewidth]{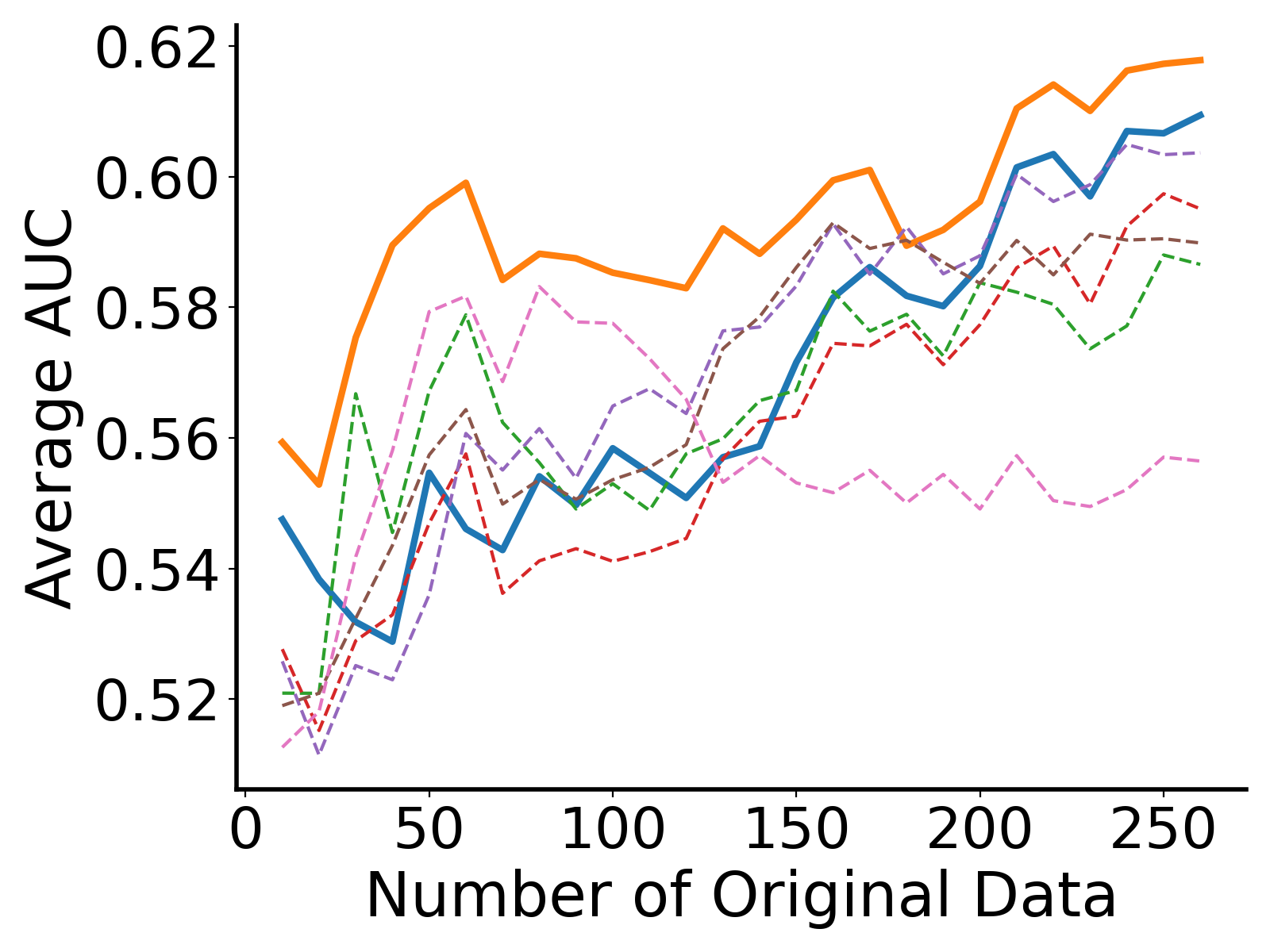}
        \caption{White male, 18-24.}
        \label{drug_0-1-6_whitemale}
    \end{subfigure}

    \caption{Average AUC for two demographic groups on the Drug dataset.}
    \label{fig:Total_Number_of_Problems}
\end{figure}

To assess model performance across diverse populations, we partition the Drug dataset into two subgroups (Drug-A and Drug-B) based on their demographic features (age, gender, and ethnicity). Figure \ref{drug_1-0-6_whitefemale} corresponds to the Drug-A subgroup with age between 25 and 34, female, and White, while Figure \ref{drug_0-1-6_whitemale} focuses on the Drug-B subgroup, i.e., White males with age between 18 and 24. For the Drug-B subgroup, T$^2$ delivers the highest AUC across all sample sizes from 10 to 250. The larger improvement of T$^2$ in Drug-B is attributed to binary label imbalance, where 74.07\% of samples are labeled as 1, compared to 63.11\% in the Drug-A subgroup. Overall, our method enhances model utility in both balanced and imbalanced settings. Table \ref{compas_all_utility} provides a summary of downstream utility across varying sizes of $D_\text{ori}$ on the COMPAS dataset. T$^2$ achieves the best overall performance across all four metrics, highlighting its ability to generate high-quality synthetic data across multiple evaluation dimensions. Detailed experimental results, including the average AUC of four ML models, are shown in Appendix~\ref{appendix_result}.

\begin{table}[htbp]
\caption{Average downstream utility (\%) on the COMPAS dataset.}
\centering
\footnotesize
\setlength{\tabcolsep}{5pt}
\renewcommand{\arraystretch}{0.8}

\begin{tabular}{@{}ccccc@{}}
\toprule
\textbf{Method} & \textbf{Accuracy} & \textbf{AUC} & \textbf{F1} & \textbf{Recall} \\
\midrule
$D_\text{ori}$ & 60.89 & 64.12 & 58.69 & 60.00 \\
SMOTE     & 60.77 & 64.86 & 58.43 & 60.20 \\
TVAE      & 60.81 & 64.32 & 58.63 & 60.62 \\
CTGAN     & 60.54 & 64.61 & 58.23 & 60.25 \\
CLLM      & 61.70 & 65.54 & 59.26 & 60.90 \\
EPIC &60.40 &63.03 &58.24 &60.79 \\
T$^2$  (\textbf{ours})                      & \textbf{62.05} & \textbf{67.29} & \textbf{61.84} & \textbf{66.11} \\
\bottomrule
\end{tabular}
\label{compas_all_utility}
\end{table}

\paragraph{Data quality.}




We evaluate the quality of generated data using three metrics: detection \citep{liu2023goggle}, $\alpha$-precision \citep{alaa2022faithful}, and $\beta$-recall \citep{alaa2022faithful}. Detection measures whether synthetic data can be distinguished from real data, with lower scores indicating higher similarity (thus better quality). $\alpha$-precision quantifies the fidelity of generated samples, while $\beta$-recall captures their diversity. Table \ref{data_qua} shows the average data quality across varying original data sizes for different methods on real-world datasets. T$^2$ consistently outperforms baseline methods except in $\alpha$-precision on COMPAS, a dataset with well-documented disparity issues \citep{wang2019empirical}. Its real distribution is highly concentrated and exhibits limited feature variability. T$^2$ is designed to generate batches that increase IG and expand cluster-level diversity, which naturally encourages exploration beyond the tight, high-probability core modes favored by $\alpha$-precision. This trade-off is beneficial for downstream learning, as reflected in stronger $\beta$-recall, detection scores, and predictive performance. These results demonstrate that T$^2$ mostly generates synthetic data that are indistinguishable from real data, faithful to the real distribution, and diverse enough to cover rare or missing modes.

\begin{table}[htbp]
\centering
\fontsize{9pt}{9pt}\selectfont
\caption{Average data quality (\%) on real-world data.}
\resizebox{\columnwidth}{!}{%
\begin{tabular}{ccccc}
\toprule
\textbf{Dataset} & \textbf{Method} & \textbf{Detection} & \textbf{$\alpha$-Precision} & \textbf{$\beta$-Recall} \\
\midrule
\multirow{7}{*}{DRUG-A}
 & SMOTE & 65.71 & 53.97 & 35.50 \\
 & TVAE & 65.59 & 53.89 & 35.73 \\
 & CTGAN & 64.60 & 54.01 & 35.77 \\
 & CLLM & 63.27 & 58.12 & 37.25 \\
 &EPIC &63.62 &65.88 &28.10 \\
 & T$^2$ (\textbf{ours}) & \textbf{56.68} & \textbf{88.76} & \textbf{48.45} \\
\midrule
\multirow{7}{*}{DRUG-B}
 & SMOTE & 66.30 & 51.20 & 34.01 \\
 & TVAE & 65.99 & 51.12 & 33.34 \\
 & CTGAN & 66.15 & 51.23 & 33.25 \\
 & CLLM & 63.67 & 56.53 & 35.47 \\
  &EPIC &66.66 & 81.63& 37.18\\
 & T$^2$ (\textbf{ours}) & \textbf{57.81} & \textbf{90.96} & \textbf{49.29} \\
\midrule
\multirow{7}{*}{COMPAS}
 & SMOTE & 52.91 & 88.47 & 46.52 \\
 & TVAE & 51.37 & 89.94 & 50.32 \\
 & CTGAN & 51.12 & 91.23 & 51.03 \\
 & CLLM & 50.26 & \textbf{91.58} & 48.63 \\
  &EPIC &53.02 &81.77 & 44.12 \\
 & T$^2$ (\textbf{ours}) & \textbf{49.17} & 91.11 & \textbf{51.29} \\
\bottomrule
\end{tabular}}
\label{data_qua}
\end{table}

\subsection{Deconstructing the Assembly Line: Ablation Studies and Sensitivity Analysis}
\begin{figure}[htbp]
  \centering
  \includegraphics[width=\columnwidth]{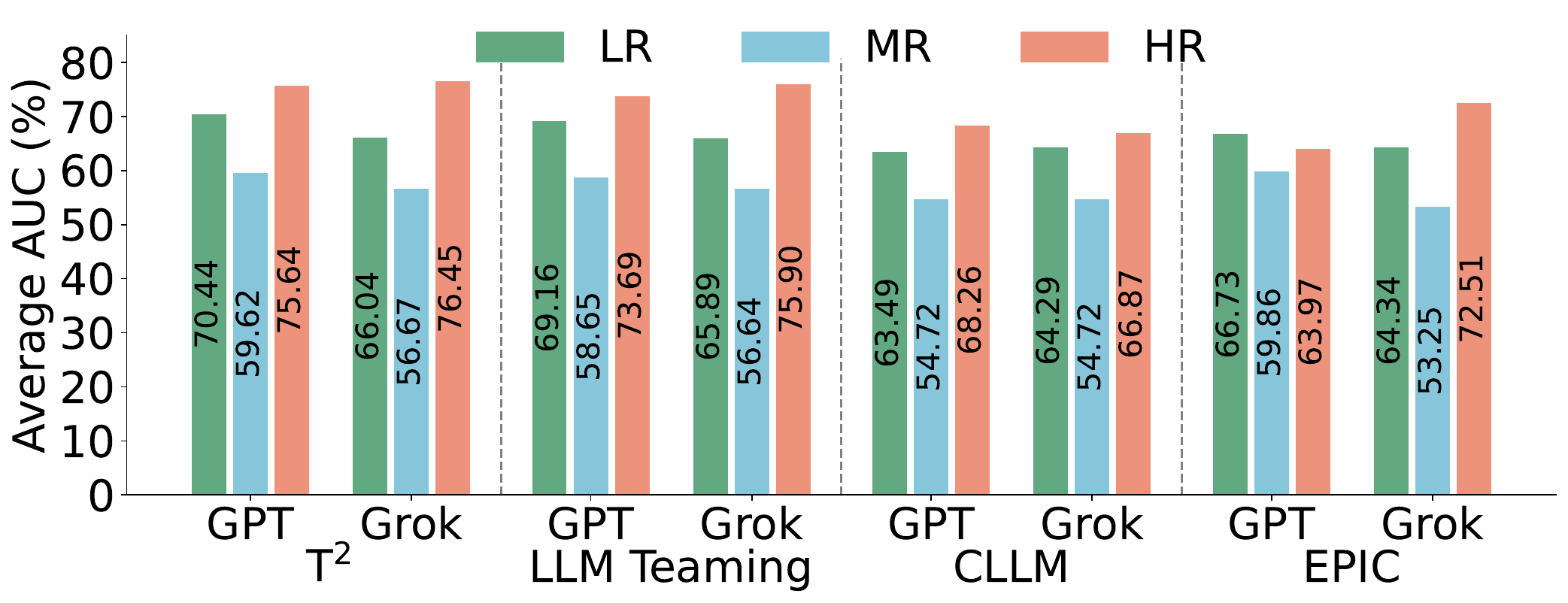}
  \caption{Average AUC of LLM backbones under data imbalance.}
  \label{data_imb_llm_impact}
\end{figure}

\paragraph{Impact of LLM backbones.}
Figure \ref{data_imb_llm_impact} analyzes how different LLM backbones affect the average AUC of LLM-based data generation under data imbalance on the Diabetes dataset. LLM teaming corresponds to T$^2$ without the QC steps. The benefits of structured LLM teaming are robust across backbones. For both GPT-5.1 and Grok-4.1-Fast, LLM teaming outperforms single-LLM baselines (CLLM and EPIC) in all three risk groups. Overall, we observe that backbone choice may influence absolute AUC levels, but does not change our conclusion that T$^2$ performs best. The performance disparities across different backbones are moderate compared to the gains from the framework choices. These results indicate that stronger LLM backbones can further amplify performance, but are not a prerequisite for the gains achieved by our framework. The robustness across Llama-3.3, GPT-5.1, and Grok-4.1-Fast supports the generality of our framework.



\begin{figure}[htbp]
  \centering
  \includegraphics[width=\columnwidth]{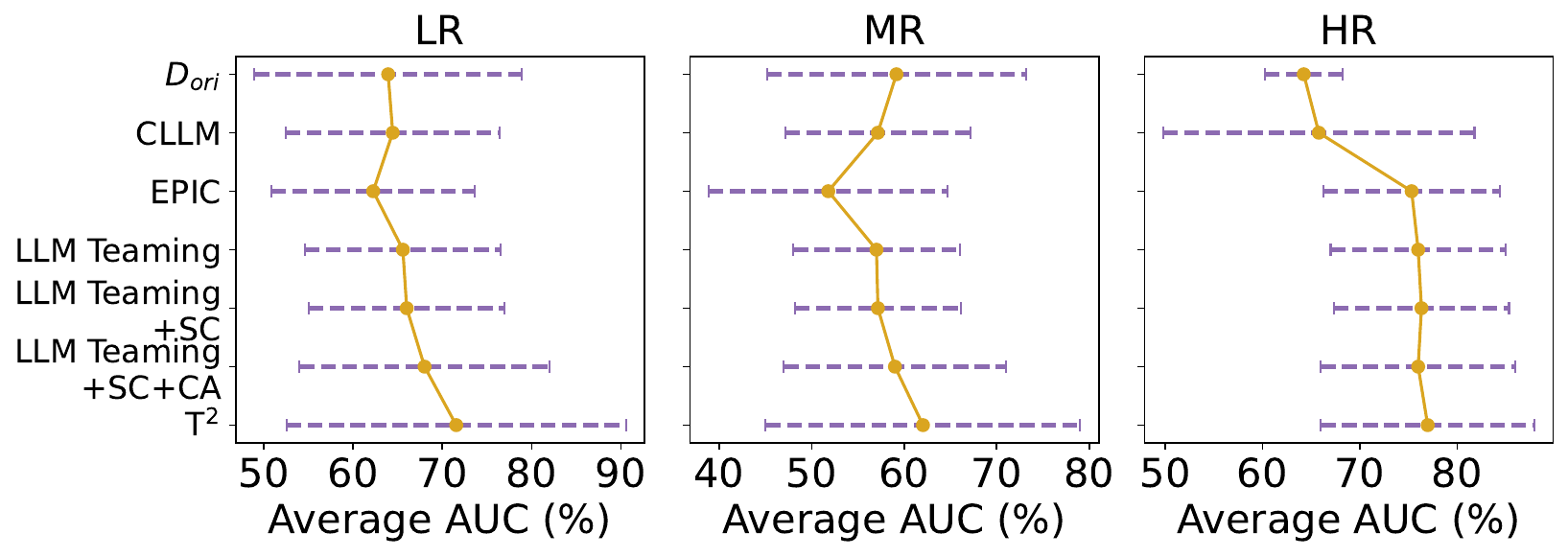}
  \caption{Average AUC with 95\% CIs of LLM teaming with individual QC stages.}
  \label{qc_contriution}
\end{figure}

\begin{figure}[htbp]
  \centering
  \begin{subfigure}{0.49\columnwidth}
    \centering
    \includegraphics[width=\linewidth]{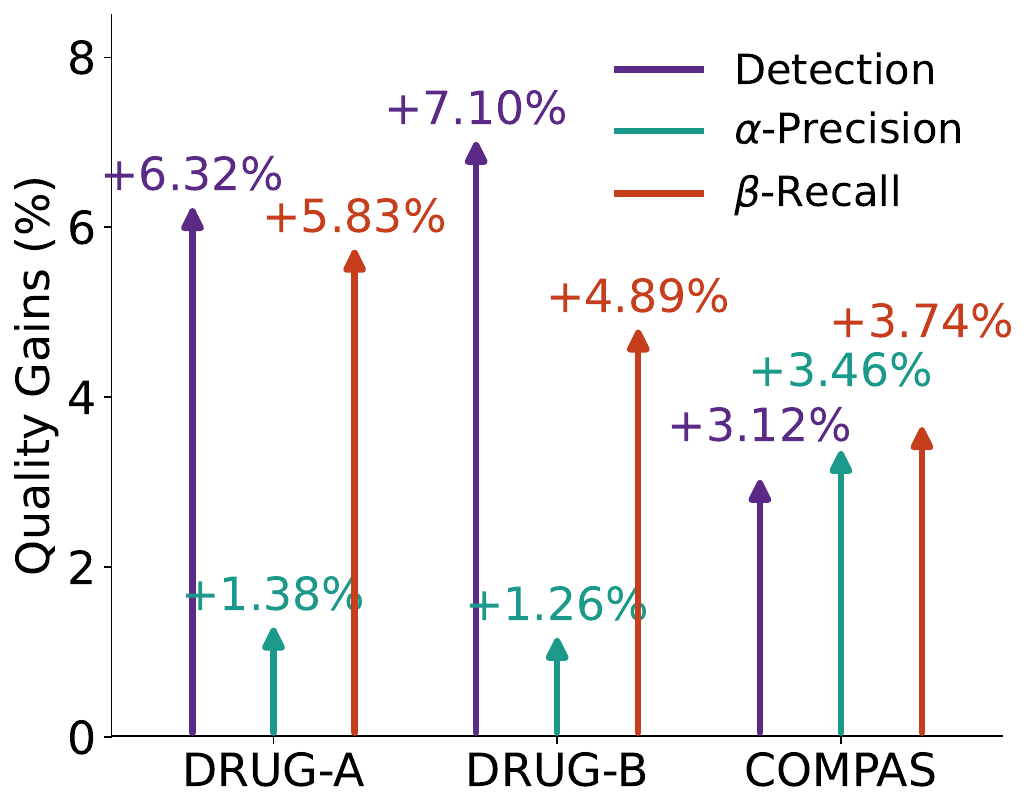}
    \caption{Data quality gains from QC over LLM teaming.}
    \label{fig:qc_quality_gains}
  \end{subfigure}\hfill
  \begin{subfigure}{0.49\columnwidth}
    \centering
    \includegraphics[width=\linewidth]{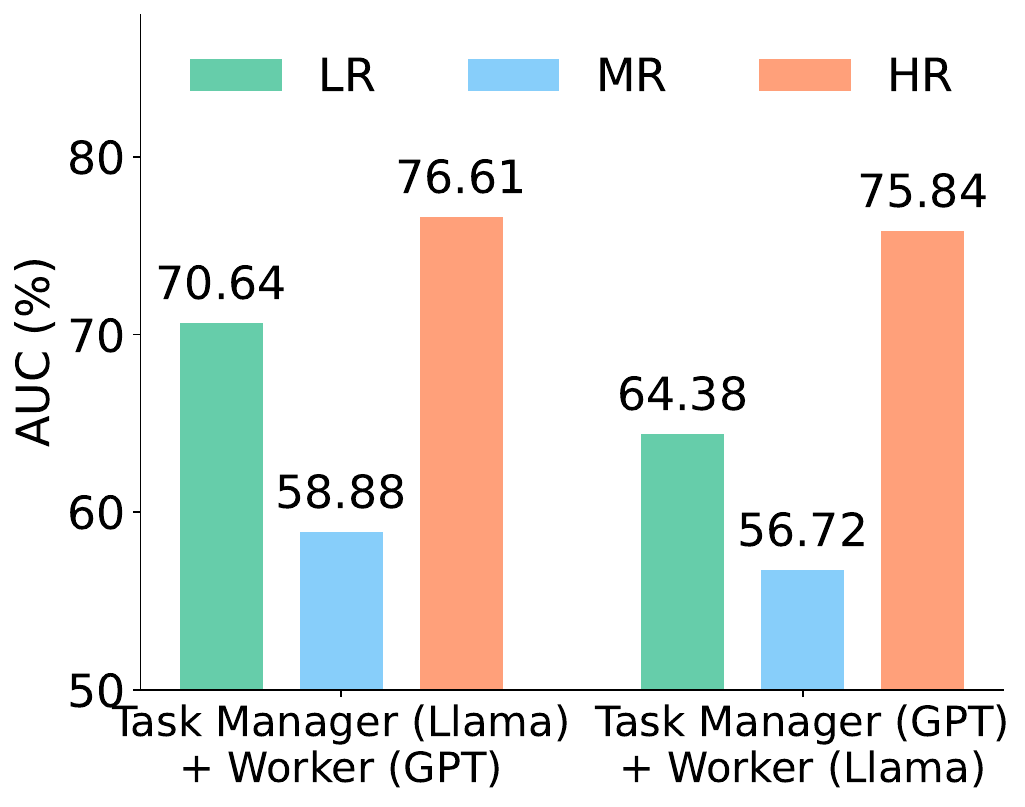}
    \caption{AUC of T$^2$ for different LLM teaming designs.}
    \label{fig:llm_teaming_design_auc}
  \end{subfigure}
  \caption{Results of QC gains and LLM teaming designs.}
  \label{fig:qc_teaming_combined}
\end{figure}

\paragraph{Contribution of QC pipeline.}
Figure \ref{data_imb_llm_impact} already demonstrates the contribution of LLM teaming. We next evaluate the contribution of the trimming components, i.e., the QC pipeline. Figure \ref{qc_contriution} decomposes the effect of each QC stage under simulated data imbalance. While LLM teaming alone already improves performance, it yields limited gains for the LR and MR groups due to the presence of noisy or weakly aligned samples. As the first QC stage, sanity check (SC) introduces only small changes, confirming that LLM teaming rarely violates validity constraints when the features are easily interpretable. The second stage of objective-related cost assessment (CA) increases AUC for LR and MR. After incorporating the final diversity inspection stage, the full T$^2$ framework achieves the highest AUC across all risk groups. The slightly wider 95\% confidence intervals (CIs) after QC reflect the trimming process, which reduces retained sample counts but improves informativeness per sample. Figure \ref{fig:qc_quality_gains} shows that QC substantially enhances data quality after LLM teaming. It yields significant gains in detection (up to +7.10\%) and $\beta$-recall (up to +5.83\%) with moderate gains in $\alpha$-precision. These results demonstrate that QC is a core component of the T$^2$ framework. LLM teaming provides a strong generative prior, while QC further enhances informativeness, diversity, and alignment.




\paragraph{Impact of LLM teaming designs.}
Figure \ref{fig:llm_teaming_design_auc} analyzes the impact of LLM teaming designs, focusing on how average AUC changes when allocating model capacity differently between the task manager and the worker using simulated imbalanced data. We view GPT-5.1 as a large model and Llama-3.3 as a smaller one, comparing two different team compositions (\textit{small manager + large worker} vs. \textit{large manager + small worker}). As shown in Figure \ref{fig:llm_teaming_design_auc}, the first configuration (left) consistently achieves a higher AUC than the inverse design (right). This indicates that allocating greater capacity to the worker yields larger gains than assigning it to the task manager. In our settings, task decomposition is relatively stable and coarse-grained (compared to statically derived causal graphs), such that a smaller LLM is generally sufficient to produce a coherent and effective generation plan. In contrast, the quality of the generated samples is largely determined by the worker LLM, where greater model capacity directly enables stronger conditional reasoning and fewer local inconsistencies. This insight provides practical guidance for cost-aware LLM teaming: when resources are limited, prioritizing stronger workers can be more beneficial than investing in a stronger task manager.




\section{Conclusion}
In this work, we introduce T$^2$, a novel assembly-line framework for tabular data generation that leverages the complementary strengths of LLMs and a principled QC pipeline. By decomposing generation into subtasks handled by specialized LLMs and trimming outputs through three-stage checks, our approach systematically transforms raw generations into high-quality synthetic data. Extensive experiments across simulated and real-world data with deficiencies (e.g., imbalance, incompleteness, noise, or scarcity) demonstrate that T$^2$ consistently outperforms state-of-the-art generation methods. Notably, T$^2$ improves downstream utility across diverse classifiers and achieves fidelity, diversity, and indistinguishability from real data. These results highlight the potential of structured LLM teaming, coupled with rigorous QC, to bridge gaps in learning under data deficiencies.

\bibliography{example_paper}

\bibliographystyle{plainnat}


\newpage
\appendix
\onecolumn

\section{Related Work}\label{appendix_sec_relatedwork}
\paragraph{Tabular data generation.}
Synthetic tabular data generation has evolved from statistical methods to deep generative models \citep{chawla2002smote, xu2019modeling, goyal2024systematic, dankar2021fake}. GOGGLE \citep{liu2023goggle} learns an explicit relational structure among columns and jointly trains a message-passing VAE, enabling better modeling of sparse, heterogeneous dependencies in tabular data. TabDiff \citep{shi2024tabdiff} proposes a unified mixed-type diffusion model that jointly handles numerical and categorical columns in continuous time, with feature-wise learnable noise schedules and a transformer denoiser. LLMs represent a paradigm shift, leveraging vast pre-trained knowledge to generate data for scenarios absent in the original dataset \citep{tang2023does, patel2024datadreamer, shimabucoro2024llm}. CLLM \citep{seedat2024curated} leverages a frozen LLM to generate tabular samples in ultra-low-data settings and then curates them via learning-dynamics signals. DataEnvGym \citep{khan2024dataenvgym} introduces a modular teacher-student testbed where data generation agents plan and synthesize training data in a feedback loop to improve a student model, framing generation as sequential decision-making. \citet{goyal2025llm} present a practical platform that fine-tunes LLMs and integrates differential privacy to generate datasets while preserving sensitive information. \citet{yang2024enhancing} define table similarity via realistic analyst-style transformations and introduces an LLM-driven pipeline that generates large-scale pairs of similar tables to train and evaluate table-level embeddings. However, single LLM can hallucinate unseen categories, violate hard constraints and struggle to capture complex conditional distributions. Our framework, T$^2$, addresses these gaps by decomposing generation into a manufacturing assembly line and integrating a rigorous three-stage QC pipeline to ensure the final dataset is plausible and task-aligned.

\paragraph{Data quality control.}
QC for synthetic data has matured from ad-hoc checks to systematic, scalable pipelines that validate schemas and repair errors \citep{baur2020quality, tamm2025data, alaa2022faithful}. \citet{schelter2018automating} support incremental metric computation on growing datasets, and add ML-assisted predictability and anomaly detection to automate large-scale data quality verification. Building on this work, \citet{schelter2019differential} propose a differential extension that represents data-quality metrics as algebraic states with commutative-monoid properties, enabling incremental, partition-aware verification without rescanning previously processed data. Recent advances increasingly emphasize QC for LLM-generated data \citep{chen2024diversity, hu2025qualityflow}. \citet{sousa2024generation} apply a human protocol to deduplicate, filter out-of-scope samples before using the curated set from LLMs and underscore the need for human validation. \citet{wang2023post} introduce a model-agnostic, differential privacy-preserving post-processing method that reweights a synthetic dataset via information projection so that selected utility measures, e.g., moments, correlations, match noisy targets from the real data. LLM-TabLogic \citep{long2025llm} uses LLM prompting to infer and compress inter-column logical relationships and then conditions a latent diffusion generator on these constraints, producing synthetic tables that better preserve logical consistency while maintaining strong fidelity and privacy. CROWDSELECT \citep{li2025crowdselect} aggregates multiple LLMs’ responses and reward scores to compute three metrics with diversity clustering and multi-metric normalization to select synthetic instruction data. However, these methods remain pointwise and model-agnostic, and do not guarantee that admitted samples are useful for the downstream task or that they cover rare modes. Our framework, T$^2$, addresses these gaps by coupling a manufacturing-style assembly line with a three-stage QC that is explicitly task-linked and batch-level.

\section{Prompt Examples}\label{app_prompt_exm}
\begin{figure}
\centering
        \includegraphics[width=1\textwidth]{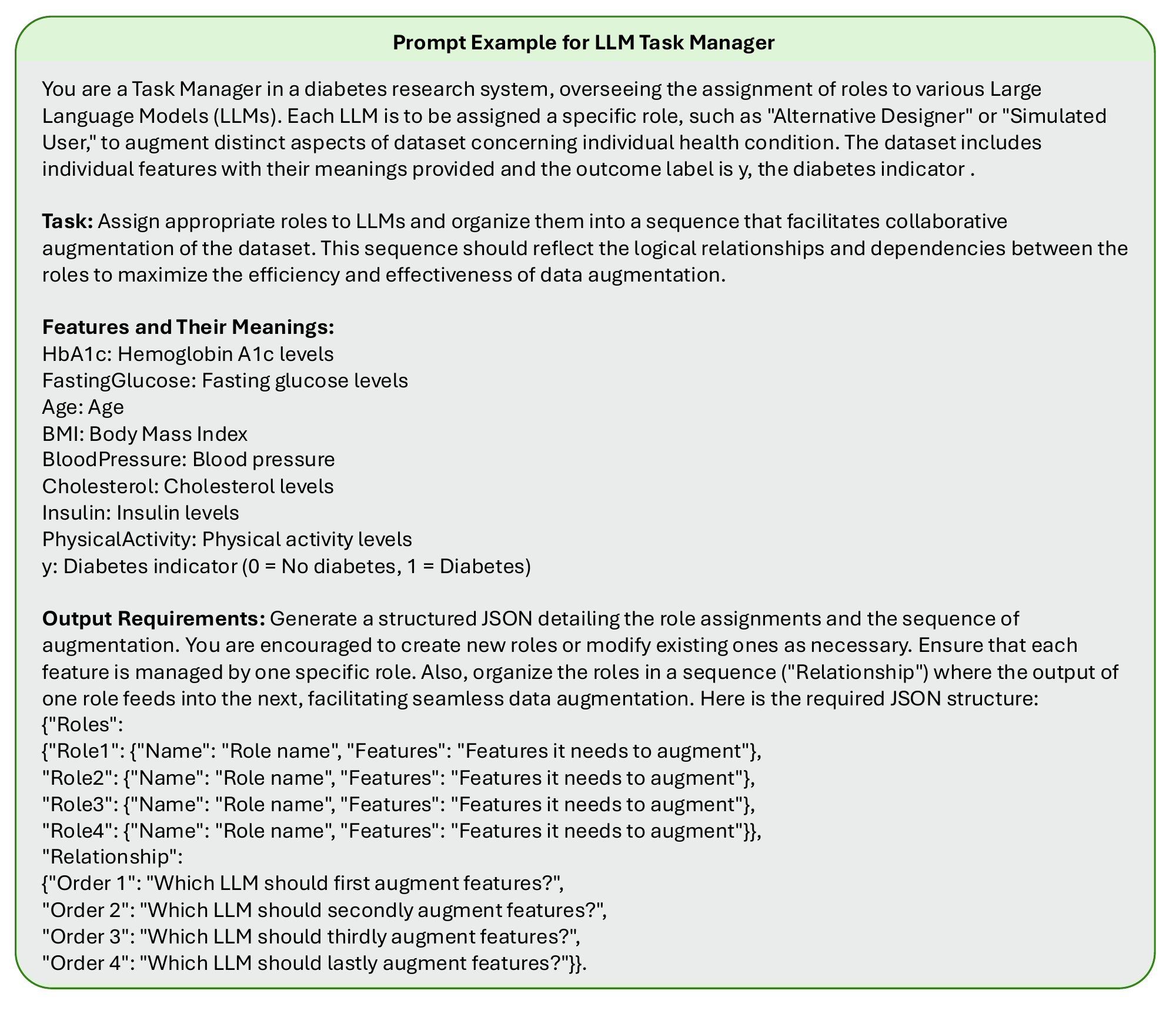}
\caption{Prompt example of task manager LLM.}\label{prompt_task_man}
\end{figure}

\begin{figure}
\centering
        \includegraphics[width=1\textwidth]{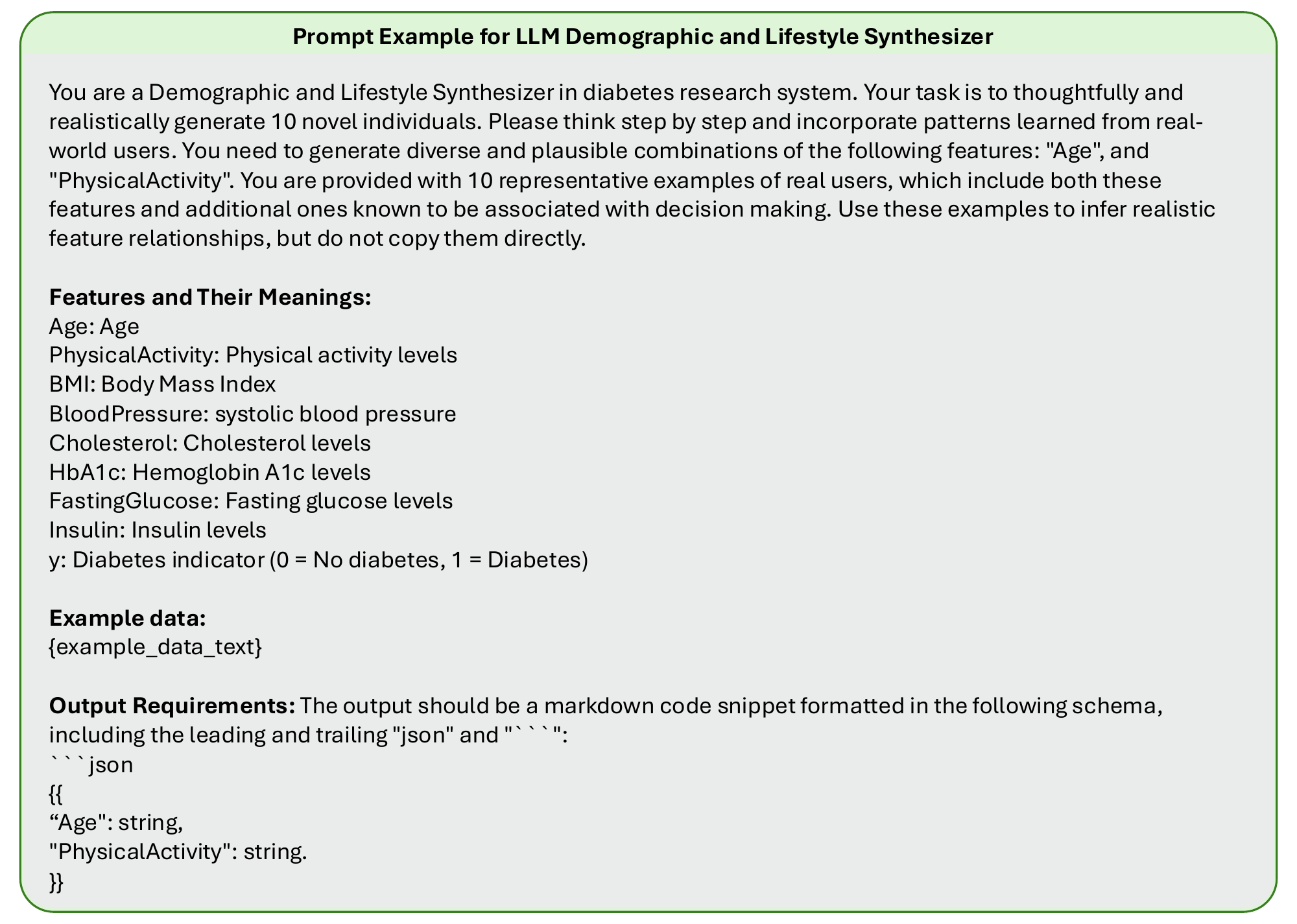}
\caption{Prompt example of an assigned role by the LLM, i.e., demographic and lifestyle synthesizer.}\label{prompt_role1}
\end{figure}

\begin{figure}
\centering

        \includegraphics[width=1\textwidth]{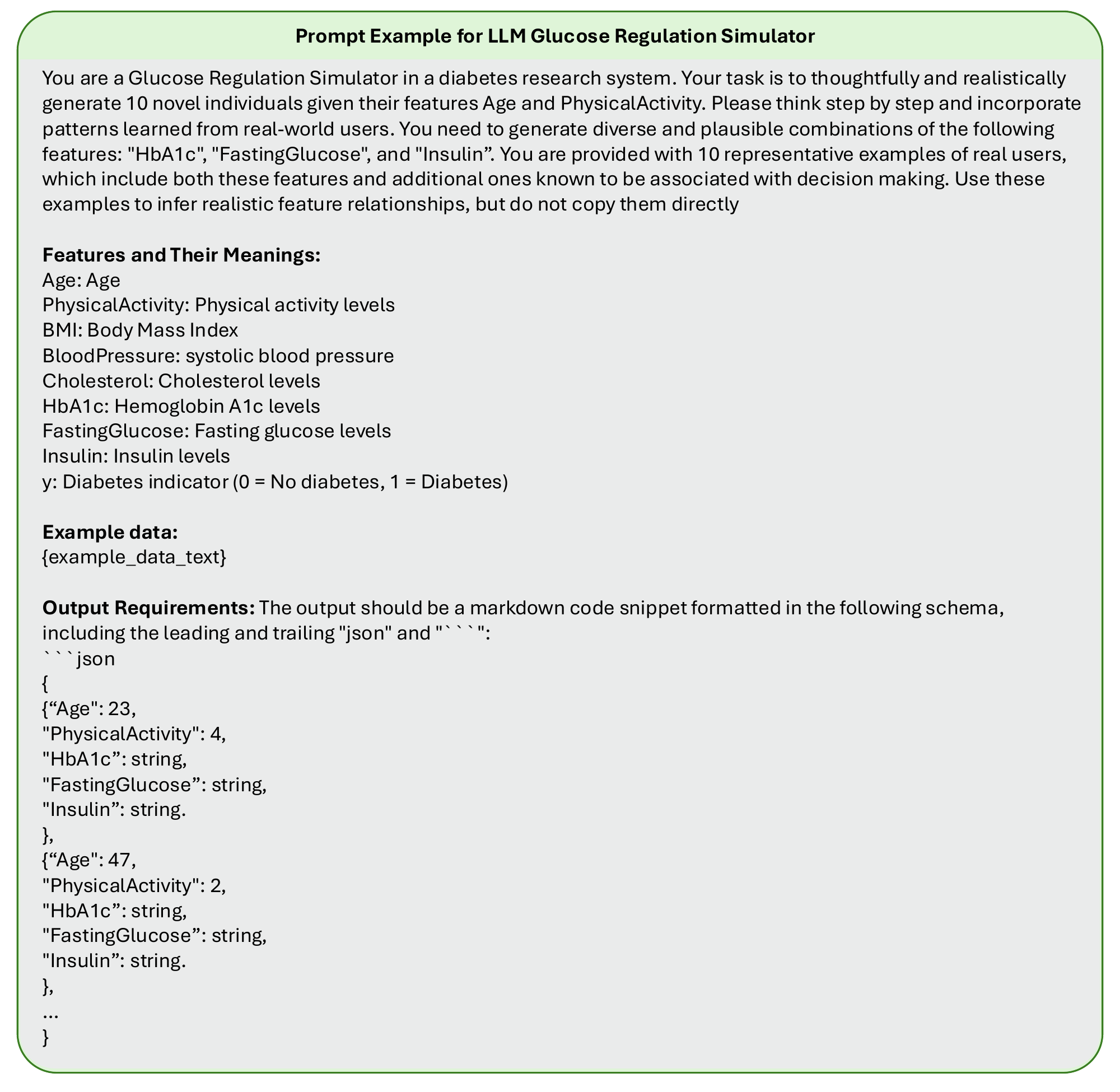}
\caption{Prompt example of an assigned role by the LLM, i.e., glucose regulation simulator, following LLM demographic and lifestyle synthesizer.}\label{prompt_role2}
\end{figure}

We provide prompt examples of the LLM task manager in Figure \ref{prompt_task_man}, along with one of its assigned roles, the LLM Demographic and Lifestyle Synthesizer, in Figure \ref{prompt_role1}, and a subsequent role, the LLM Glucose Regulation Simulator, in Figure \ref{prompt_role2}. The prompts in our framework follow a fixed and modular template that specifies task instructions, feature semantics, the original data excerpt, and output requirements, rather than relying on dataset-specific heuristics or hand-crafted prompt tuning. The task manager automatically constructs the dependency structure of components from the feature dictionary, and for different datasets or LLM workers, the varying elements in the prompt are (i) the role name assigned by the task manager, (ii) the feature names determined by the dataset, and (iii) the feature descriptions provided in the data dictionary. These pieces are inserted into the same prompt template in a systematic way, so no manual prompt redesign is needed when switching datasets. Moreover, even if minor prompt variations introduce stochastic deviations in the data generations, our three-stage QC pipeline rigorously evaluates each batch and filters out samples whose validity, objective alignment, or diversity deteriorates due to prompting noise. As a result, the framework remains stable and effective without relying on dataset-specific prompt engineering.


\section{Experimental Details}\label{exp_detail}
\begin{figure}[htbp]
\centering
\makebox[\textwidth]{%
        \includegraphics[width=0.85\textwidth]{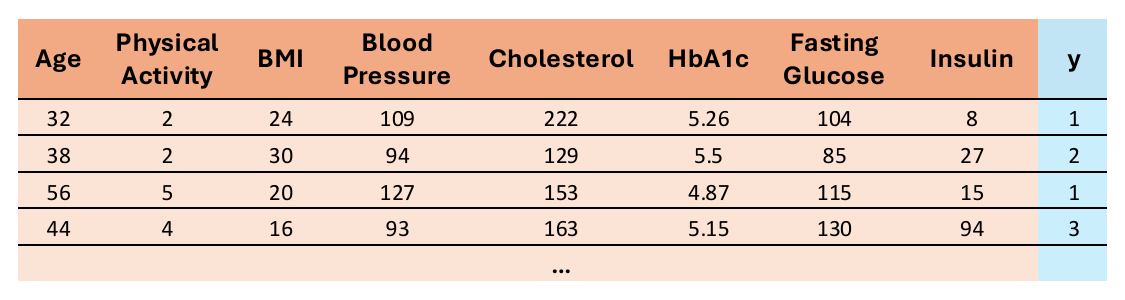}
    }
\caption{Examples of original data in Diabetes dataset.}\label{example_ori_data_diabetes}
\end{figure}

\paragraph{Datasets simulation.}
In the simulated studies, the features of the Diabetes dataset and their coefficients are generated following the distributions in \citet{rauba2024self}, with label probabilities computed using the sigmoid function. Figure \ref{example_ori_data_diabetes} shows examples of original data in the simulated Diabetes. The dataset includes eight features and one outcome label $y$. For the TravelBehavior dataset, features are simulated according to the distributions in \citet{feng2020modeling} and \citet{lin2024modeling}, and ground-truth labels are assigned using their proposed Latent Decision Threshold model. The simulated TravelBehavior dataset consists of four features and one binary outcome label. In both datasets, we construct a test set of 500 samples, while the size of the original training data varies from 10 to 100, increasing in steps of 10. Unless otherwise specified, the data generation adheres to the aforementioned distributions. We use symmetric label flipping in all noise experiments. Specifically, for a given flip ratio, we uniformly sample a subset of data points without conditioning on class and randomly flip their labels. No class receives preferential or targeted noise.

\paragraph{Real-world datasets.} 
For the real-world applications, the Drug dataset from the UCI repository \citep{fehrman2017five} is partitioned by age, gender, and ethnicity, and we only retain subgroups with more than 250 observations to ensure sufficient training and test data. After partition, the dataset consists of 24 features and one binary label outcome. For each subgroup, 100 samples are held out as the test set, with the remainder used for training after stratifying by label distribution. In the COMPAS dataset from OpenML \citep{angwin2016machine}, the number of training data varies across experiments with balanced labels, and 500 samples are reserved for testing. The COMPAS dataset consists of 13 features and one binary label outcome.

\paragraph{Data generation.}
Considering plausibility of evaluation and the token limits, we generate 10 data points in each batch for Diabetes and Drug datasets, and 20 data points in each batch for the TravelBehavior and COMPAS datasets. In the data imbalance and incompleteness experiments, the number of generated batches before trimming is fixed at 10. In all other experiments with varying original data, it is set equal to the size of the original dataset; for example, with 10 original samples, one batch is generated. No additional batches are added after trimming. This design ensures that the amount of synthetic data scales proportionally with the available data, preventing over-generation.

\paragraph{QC.} 
In step 2 objective-related cost assessment of QC pipeline, we choose $\beta=0.1$. To calculate the threshold $\tau$, we run 1000 times to obtain the mean and standard deviation. The coefficient of standard deviation ranging from 0 to 3 is selected by 5-fold cross validation. In step 3 diversity-related monitoring, we choose the number of clusters from $\{3, 4, 5\}$ to maximize the silhouette scores. The entropy improvement threshold is set as $30\%$.

\paragraph{Models.} 
All the downstream models are implemented by scikit-learn \citep{scikit-learn}. All the experiments are run on Apple M3 Pro. We use SynthCity \citep{qian2023synthcity} to implement the baseline methods and to evaluate the $\alpha$-precision and $\beta$-recall, except for CLLM, which follows the implementation details in \citet{seedat2024curated}. The detection score is computed following the procedure in \citet{liu2023goggle}.

\paragraph{Evaluation metrics.} 
We use various evaluation metrics to capture different aspects of downstream predictive performance and quality of generated data. AUC measures the probability that a randomly chosen positive sample receives a higher predicted score than a randomly chosen negative sample. AUC is threshold-independent and particularly informative in imbalanced settings, which is why it serves as one of our primary metrics. Accuracy provides a holistic overview of performance but may be less informative under label imbalance. Recall measures the fraction of positive samples that are correctly identified. F1-score balances the trade-off between precision and recall and serves as a compact summary of classifier performance when neither error mode is dominant.

\section{Additional Results}\label{appendix_result}
\paragraph{LLM workers.} Task manager LLM assigns different roles for different datasets. For diabetes dataset, the LLM roles are Demographic and Lifestyle Synthesizer, Glucose Regulation Simulator, Cardiometabolic Generator, and Diabetes Expert. For TravelBehavior dataset, the LLM roles are Alternative Designer, Incentive Allocator, and Decision Predictor. For Drug dataset, the LLM roles are Demographic Profiler, Personality Synthesizer, Drug Usage Generator, and Nicotine Propensity Estimator. For COMPAS dataset, the LLM roles are Demographic Profile Designer, Juvenile History Synthesizer, Criminal Record Analyst and Recidivism Outcome Evaluator.

\paragraph{Data incompleteness.}

\begin{figure}[htbp]
\centering
    \begin{subfigure}{0.35\textwidth} 
        \centering
        \includegraphics[width=\linewidth]{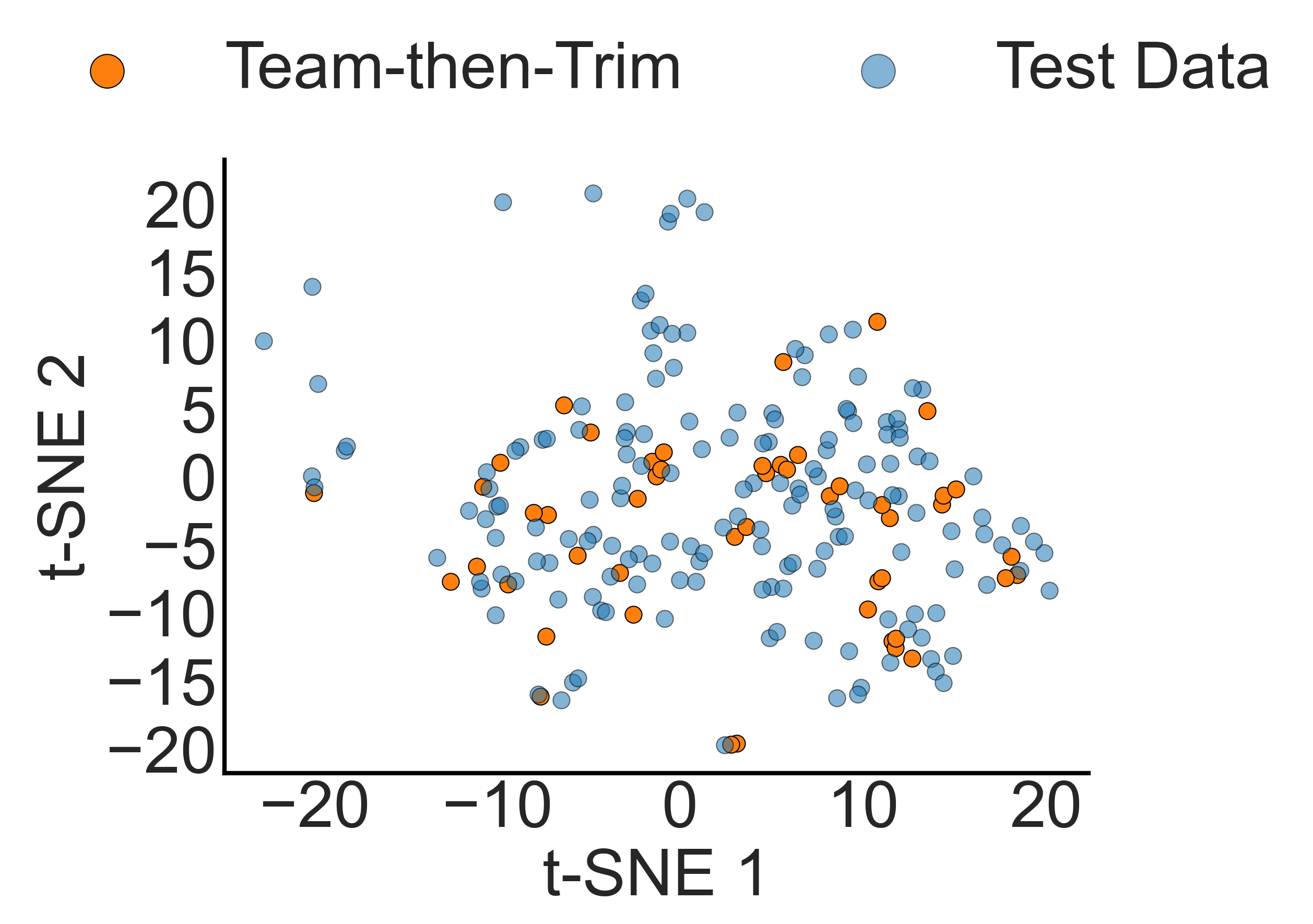}
        \caption{MR group.}
        \label{tsne_8_2_0_MR}
    \end{subfigure}%
    \begin{subfigure}{0.35\textwidth} 
        \centering
        \includegraphics[width=\linewidth]{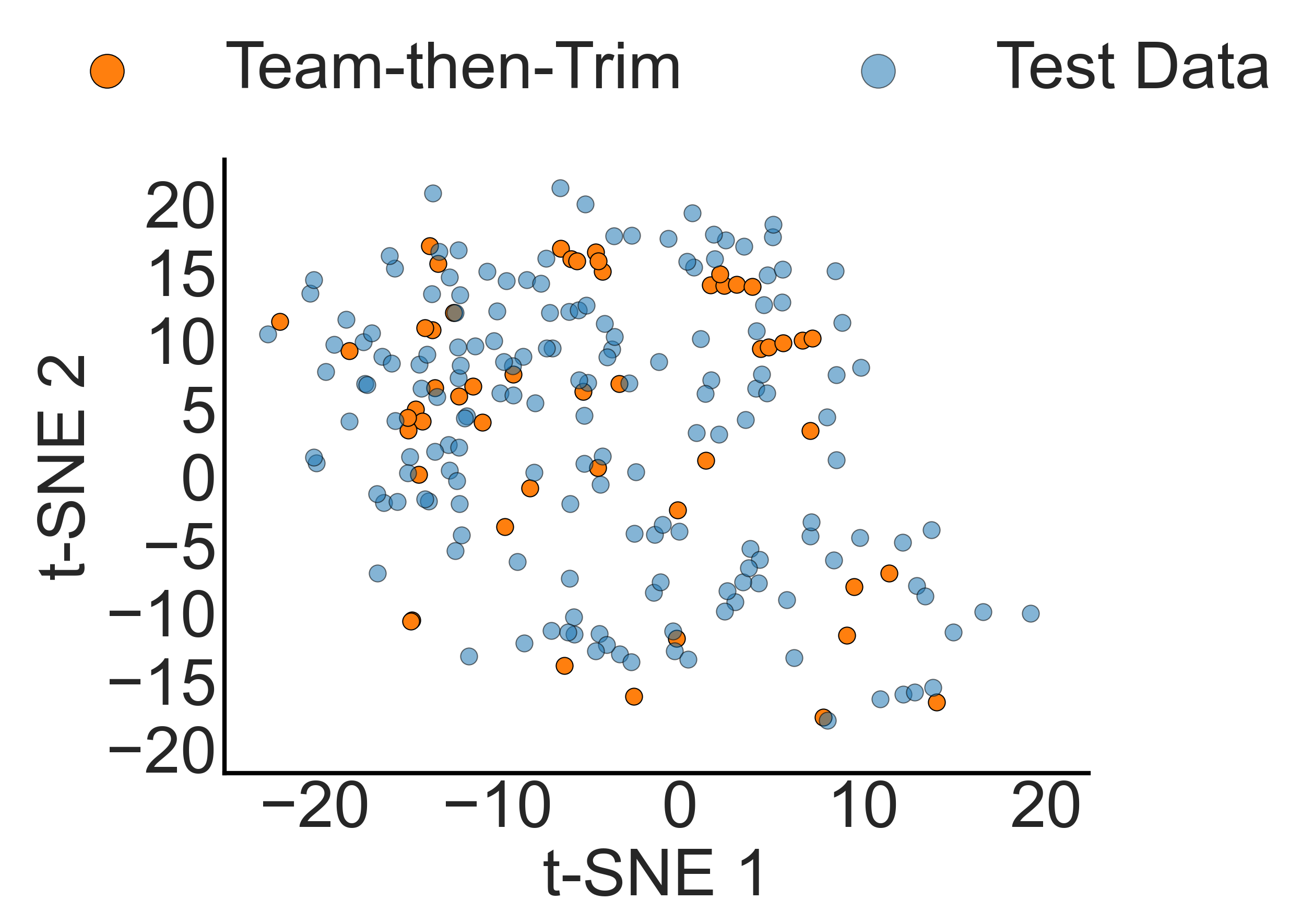}        
        \caption{HR group.}
        \label{tsne_8_2_0_HR}
    \end{subfigure}%
    \caption{t-SNE plots of MR and HR groups when the dataset is incomplete.}
    \label{tsne_8_2_0}
\end{figure}

We visualize the t-SNE embeddings of the MR and HR groups in Figure \ref{tsne_8_2_0} given the training set missing the HR subgroup on diabetes data. In both groups, the T$^2$ samples closely intermingle with the test samples, indicating that the generated data preserves the manifold structure of the real-world distribution. This alignment reflects the ability of our LLM teaming generation, followed by QC, to enrich existing but underrepresented or even missing subpopulations with high fidelity. The results show that, unlike classical baselines, T$^2$ can generalize beyond observed data to reconstruct missing modes in the distribution. This capability enables downstream models to achieve strong performance on previously unseen subgroups.

\paragraph{Data noise.}
Table \ref{diabetes_noisy_other_metric} reports the average performance on the diabetes dataset under varying levels of label noise, with flip ratios of 0.2, 0.3, and 0.4. Under moderate and high label corruption, T$^2$ can achieve stronger performance compared with when flip ratio=0.2, delivering the most consistent improvements. The gains over $D_\text{ori}$ and all baselines highlight the effectiveness of combining LLM teaming with rigorous trimming in filtering out mislabeled or uninformative samples. At a low noise level (flip ratio = 0.2), all methods achieve relatively high AUC, F1 scores and recall, reflecting the mild corruption. T$^2$ achieves the best AUC of 81.42\% and competitive F1 and recall. This indicates that even when data are only lightly corrupted, our framework provides additional robustness without overfitting to mislabeled samples.

\begin{table}[t]
\centering
\caption{Average performance (\%) across different sizes of $D_\text{ori}$ (from 10 to 100) when data is noisy on the Diabetes data.}
\label{baseline_intro}
\begin{tabular}{c c c c c}
 \toprule
 \textbf{Flip Ratio} & \textbf{Method}  & \textbf{AUC} & \textbf{F1} & \textbf{Recall} \\ 
  \hline

\multirow{8}{*}{0.2} 
   & $D_\text{ori}$ & 79.00 & \textbf{73.26} & 75.13\\
   & SMOTE     & 78.73 & 70.81 & \textbf{75.90}\\ 
   & TVAE      & 77.47 & 72.32 & 75.35 \\
   & CTGAN     & 75.38 & 72.99 & 74.07 \\
   & CLLM      & 77.99 & 68.70 & 68.91\\
     &EPIC &75.19 & 72.65 &72.23 \\
   & T$^2$ (\textbf{ours})  & \textbf{81.42} & 72.62 & 74.84\\ 
  \hline

\multirow{8}{*}{0.3}
   & $D_\text{ori}$ & 64.42 & 64.66 & 67.41 \\
   & SMOTE     & 69.77 & 63.03 & 69.57\\ 
   & TVAE      & 63.67 & 64.76 & \textbf{70.21}\\
   & CTGAN     & 66.44 & 62.87 & 68.65\\
   & CLLM      & 69.96 & 61.16 & 61.88 \\
     &EPIC &70.23 &64.37  &64.53\\
   & T$^2$ (\textbf{ours})  & \textbf{71.84} & \textbf{64.78} & 67.69  \\ 
  \hline

\multirow{8}{*}{0.4}
   & $D_\text{ori}$ & 59.18 & 58.34 & 58.91 \\
   & SMOTE     & 62.66 & 59.17 & 59.53\\ 
   & TVAE      & 61.52 & 56.72 & 57.82\\
   & CTGAN     & 57.40 & 54.20 & 55.23\\
   & CLLM      & 63.86 & \textbf{60.64} & 49.73 \\
     &EPIC &57.51 &57.88 &59.49 \\
   & T$^2$ (\textbf{ours})  & \textbf{65.94} & 58.95 & \textbf{59.76}  \\ 
 \bottomrule
\end{tabular}\label{diabetes_noisy_other_metric}
\end{table}

\paragraph{Computational costs of LLM-based data generation.}
To contextualize the computational costs of our framework, we examine the token usage and runtime of each LLM component during data generation under the diabetes data-imbalance setting. Because the QC stages rely on lightweight procedures and small models relative to our data size, its computational overhead is negligible compared with the cost of LLM-based generation. The final generated dataset has a dimensionality of 10 rows $\times$ 9 columns. As shown in Table \ref{comp_cost}, T$^2$ incurs moderately higher token consumption and runtime than CLLM; however, the framework remains cost-efficient given its substantially stronger downstream utility. Specifically, compared with AUC of $D_\text{ori}$, T$^2$ improves AUC by 7.63\% in the LR group, 2.87\% in the MR group, and 12.74\% in the HR group. These gains are consistently higher than those of CLLM, highlighting that structured LLM teaming does not introduce prohibitive overhead and, in fact, achieves a highly favorable cost–benefit tradeoff in data-deficiency regimes.

\begin{table}[htbp]
\centering
\fontsize{9pt}{9pt}\selectfont
\caption{Computational costs and AUC gains over $D_\text{ori}$ under the simulated imbalance setting on Diabetes.}
\begin{tabular}{cccccccc}
\toprule
\multirow{2}{*}{\textbf{Method}} & \multirow{2}{*}{\textbf{Component}} & \multicolumn{3}{c}{\textbf{Computational Costs}} & \multicolumn{3}{c}{\textbf{Average AUC Gains (\%)}}\\
\cmidrule(lr){3-5} \cmidrule(lr){6-8}
 &  & \textbf{Tokens (Input)} & \textbf{Tokens (Output)} & \textbf{Time (s)} & \textbf{LR} & \textbf{MR} & \textbf{HR}\\
\midrule

CLLM & -- & 771 & 872 & 6.56 & +0.51 & -2.01 & +1.55 \\
\hdashline
\multirow{4}{*}{T$^2$ (\textbf{ours})}
  & Worker LLM 1 & 832 & 387 & 4.83 & \multirow{4}{*}{\textbf{+7.63}} & \multirow{4}{*}{\textbf{+2.87}} & \multirow{4}{*}{\textbf{+12.74}} \\
  & Worker LLM 2 & 1831 & 887 & 6.08 &  &  & \\
  & Worker LLM 3 & 2457 & 1483 & 7.16 &  &  & \\
  & Worker LLM 4 & 2533 & 1507 & 7.46 &  &  & \\


\bottomrule
\end{tabular}
\label{comp_cost}
\end{table}

\begin{figure}[htbp]
    \centering
    \begin{subfigure}{0.32\textwidth} 
        \centering
        \includegraphics[width=\linewidth]{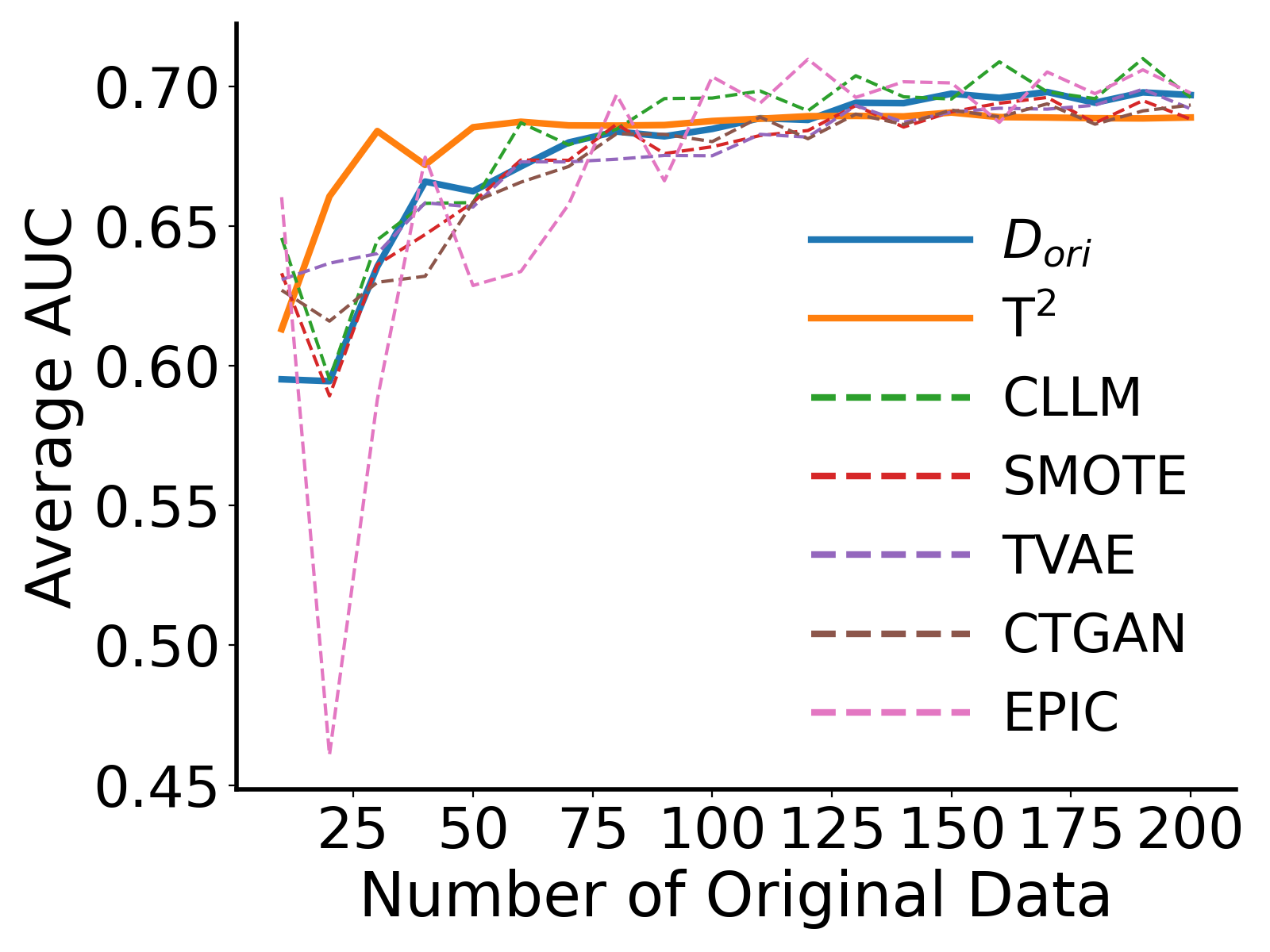}
        \caption{Logistic regression.}
        \label{fig:sub_offensiveness}
    \end{subfigure}%
    \begin{subfigure}{0.32\textwidth} 
        \centering
        \includegraphics[width=\linewidth]{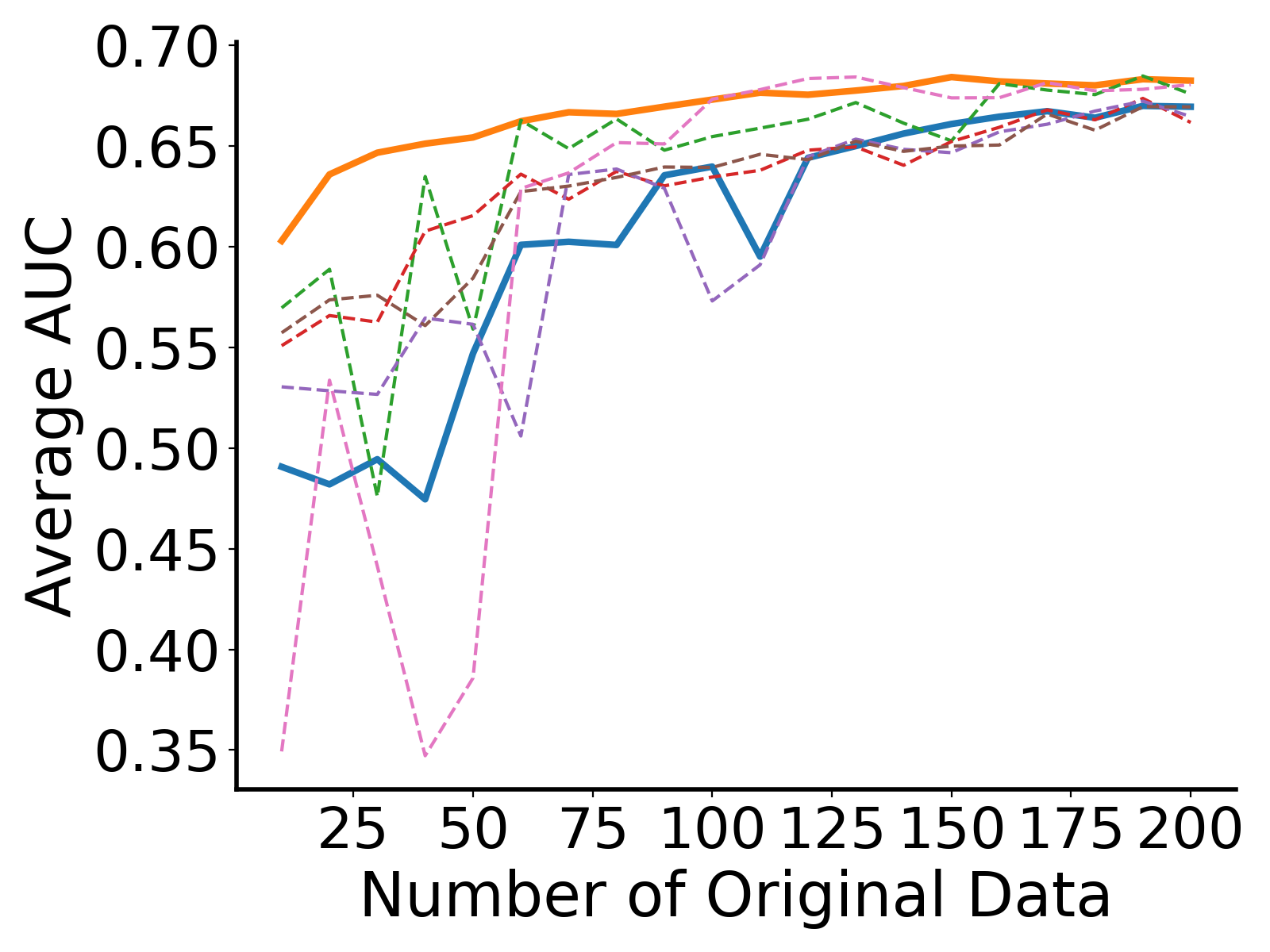}
        \caption{SVM.}
        \label{fig:sub_politeness}
    \end{subfigure}%
    \\
    \begin{subfigure}{0.32\textwidth} 
        \centering
        \includegraphics[width=\linewidth]{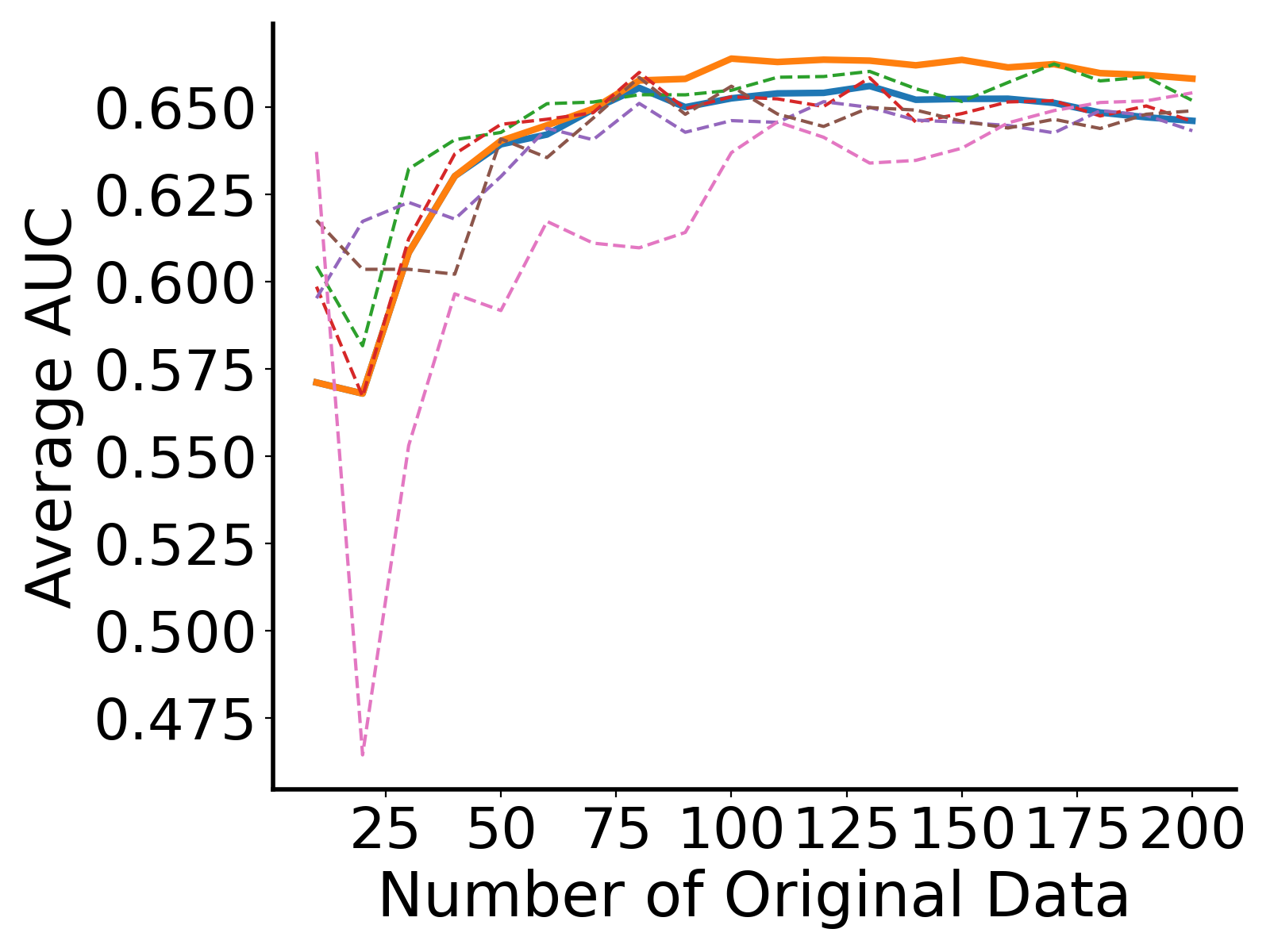}
        \caption{Random forest.}
        \label{fig:sub_qa_difficulty_rf}
    \end{subfigure}%
    \begin{subfigure}{0.32\textwidth} 
        \centering
        \includegraphics[width=\linewidth]{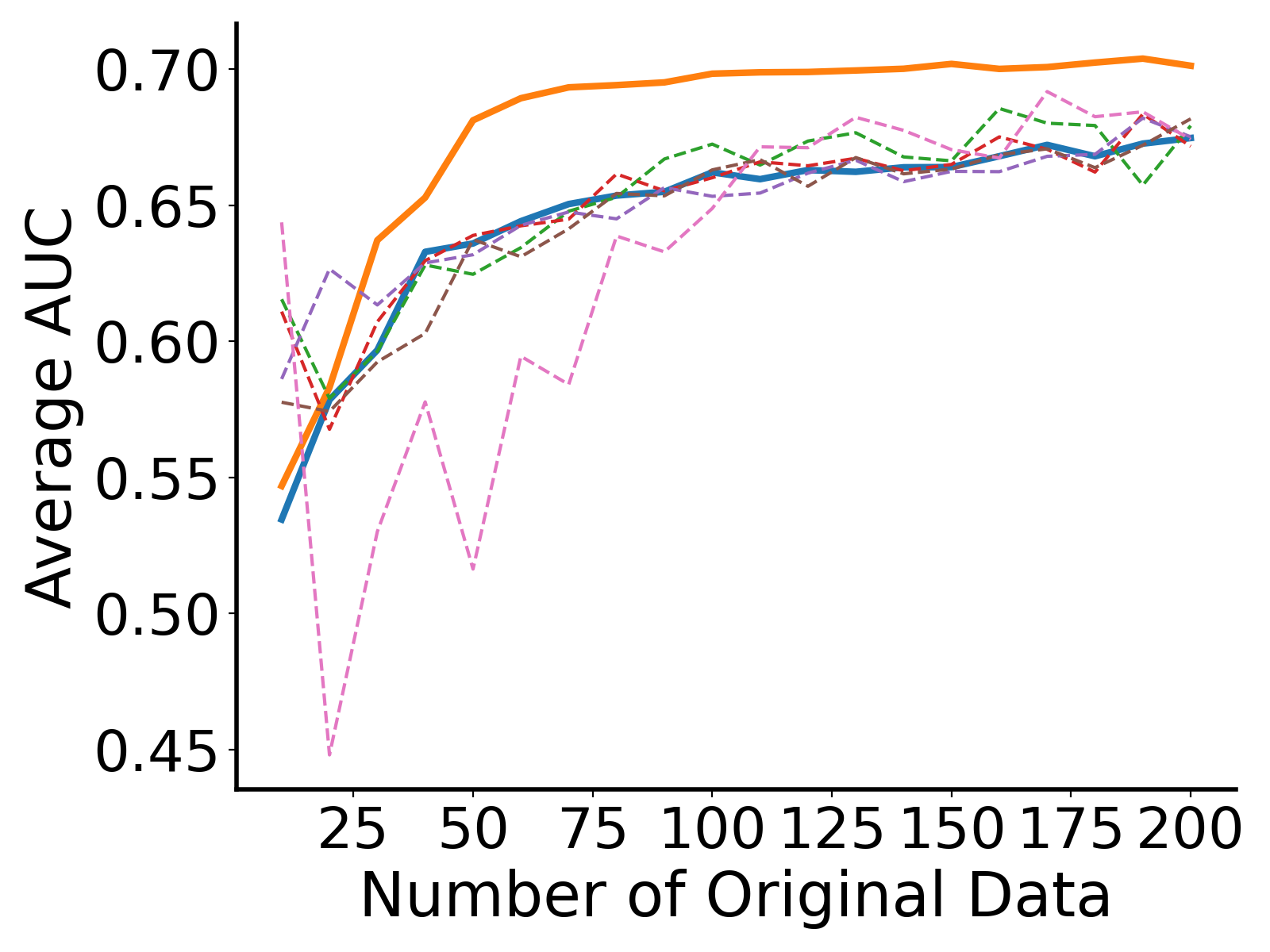}
        \caption{MLP.}
        \label{fig:sub_qa_difficulty}
    \end{subfigure}%
    \caption{Average AUC of four ML models across original data from 10 to 200 on the COMPAS dataset.}
    \label{compas_AUC}
\end{figure}

\paragraph{Model utility.}
Figure \ref{compas_AUC} presents the average AUC of four ML models on the COMPAS dataset across varying sizes of original data from 10 to 200. Across all models, T$^2$ outperforms baseline approaches under most sizes of original data. Importantly, while classical baselines occasionally achieve moderate gains, their improvements are inconsistent and often fall short of the stability achieved by T$^2$. These trends indicate that the coordinated LLM teaming plus trimming pipeline enhances model utility in ways that benefit a broad spectrum of downstream learners.

\begin{figure}[htbp]
  \centering
  \includegraphics[width=0.55\columnwidth]{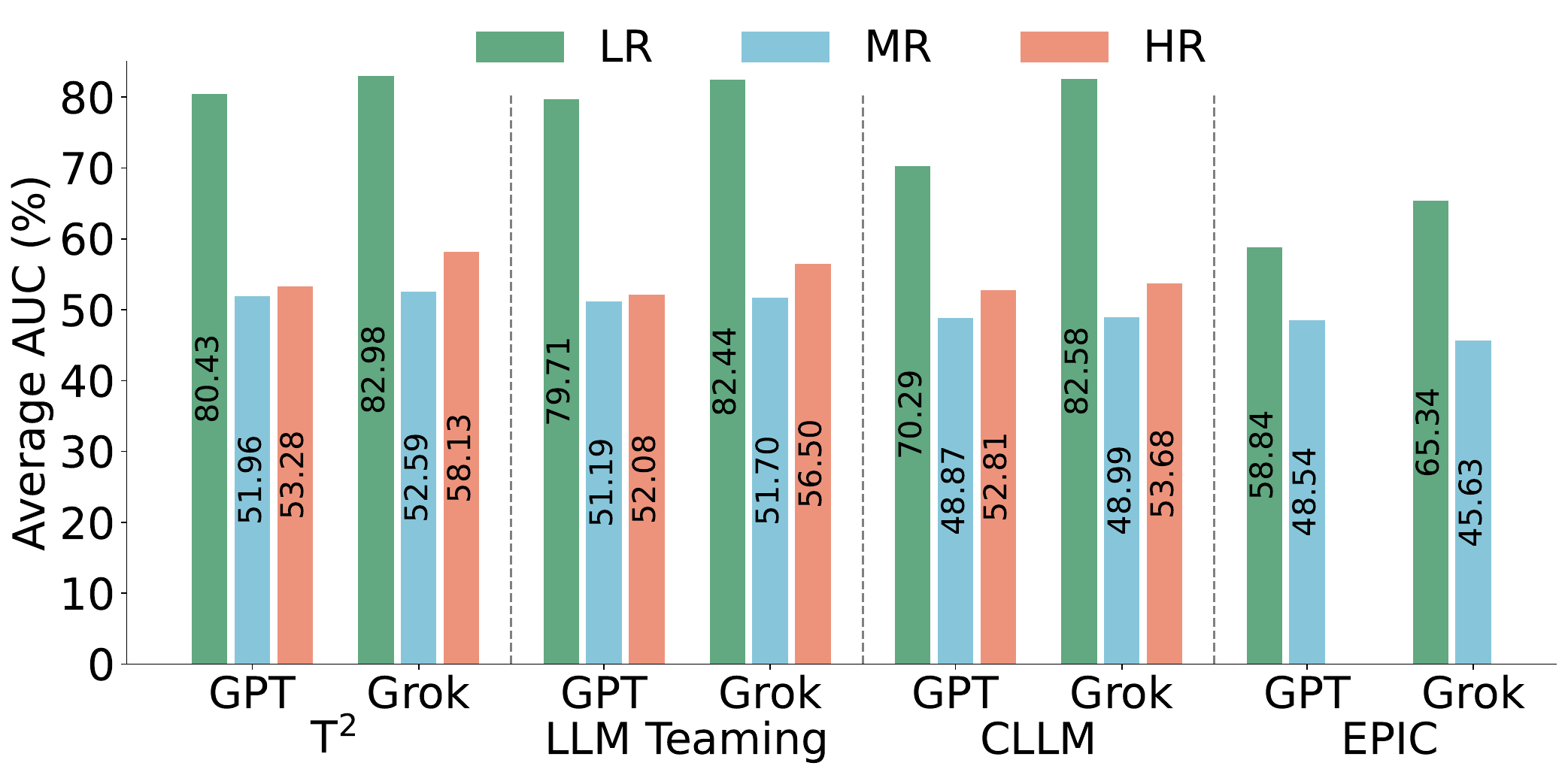}
  \caption{Average AUC of different LLM backbones under data incompleteness.}
  \label{data_inc_llm_impact}
\end{figure}

\paragraph{Impact of LLM backbones.}
Figure \ref{data_inc_llm_impact} analyzes how different LLM backbones affect the average AUC of LLM-based data generation under data incompleteness on the Diabetes dataset. The trends mirror those in Figure \ref{data_imb_llm_impact}, indicating that stronger LLM backbones can further amplify performance but are not required for the gains achieved by T$^2$.

\begin{figure}[htbp]
    \centering
    \begin{subfigure}{0.4\columnwidth}
        \centering
        \includegraphics[width=\linewidth]{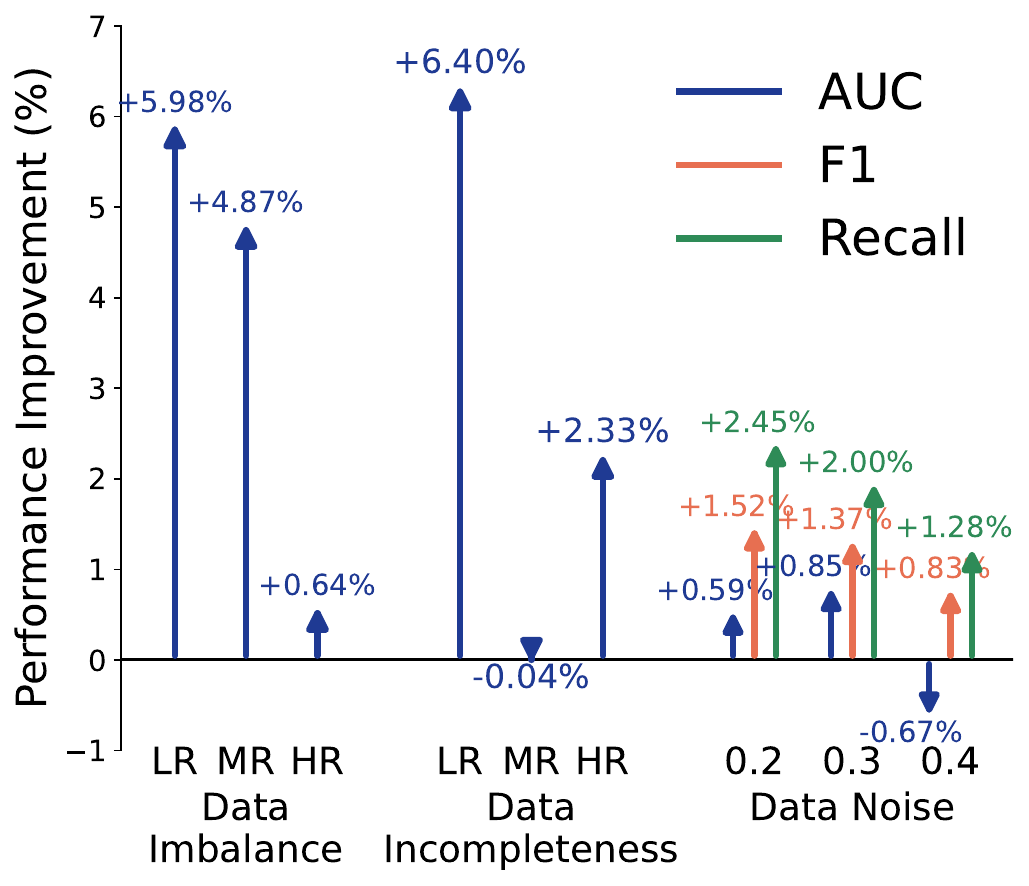}
        \caption{Simulated diabetes data.}
        \label{qc_improvement_all}
    \end{subfigure}
    \begin{subfigure}{0.4\columnwidth}
        \centering
        \includegraphics[width=\linewidth]{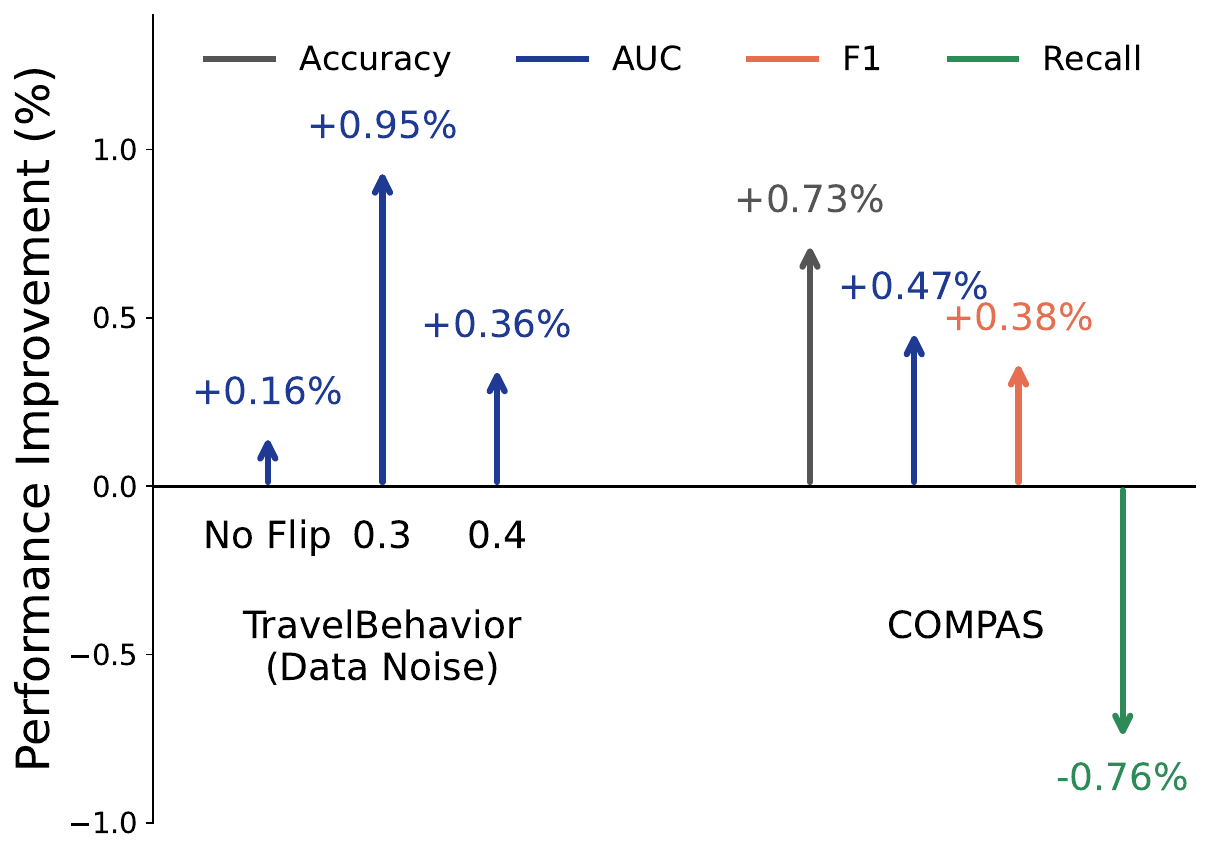}
        \caption{Simulated TravelBehavior and real-world COMPAS.}
        \label{QC_contribution_travel_compas}
    \end{subfigure}

    \caption{Downstream utility gains from QC by comparing T$^2$ with LLM teaming across different settings.}
\end{figure}

\paragraph{Contribution of QC pipeline.}
Figure \ref{qc_improvement_all} further illustrates the performance gains introduced by QC by comparing T$^2$ with LLM teaming across different data deficiencies on the Diabetes dataset. QC delivers AUC gains of +4.87\% for MR under imbalance and +2.33\% for HR under incompleteness, demonstrating its effectiveness in recovering rare or missing subpopulations. Under label noise, QC continues to improve F1 and recall across all flip ratios, even when AUC gains become marginal. This highlights that QC does not merely optimize a single metric, but instead promotes robustness by balancing precision-recall trade-offs in noisy regimes. Figure \ref{QC_contribution_travel_compas} quantifies the downstream utility gains introduced by the QC pipeline by comparing the full T$^2$ framework with LLM teaming on the TravelBehavior and COMPAS datasets. On TravelBehavior, QC consistently improves AUC across all noise levels, with gains of +0.16\% under no flip and increasing to +0.95\% and +0.36\% under flip ratios of 0.3 and 0.4, respectively. These improvements indicate that QC effectively filters out noise-amplifying samples and enhances robustness under label corruption. On the COMPAS dataset, QC yields consistent gains in accuracy, AUC, and F1, demonstrating improved overall predictive utility. Although recall slightly decreases, this trade-off reflects the objective-aligned trimming behavior of QC. Overall, these results show that QC contributes meaningful and complementary improvements beyond LLM teaming by refining raw generations into samples that are more informative.

\begin{figure}[htbp]
\centering
    \begin{subfigure}{0.49\textwidth} 
        \centering
        \includegraphics[width=\linewidth]{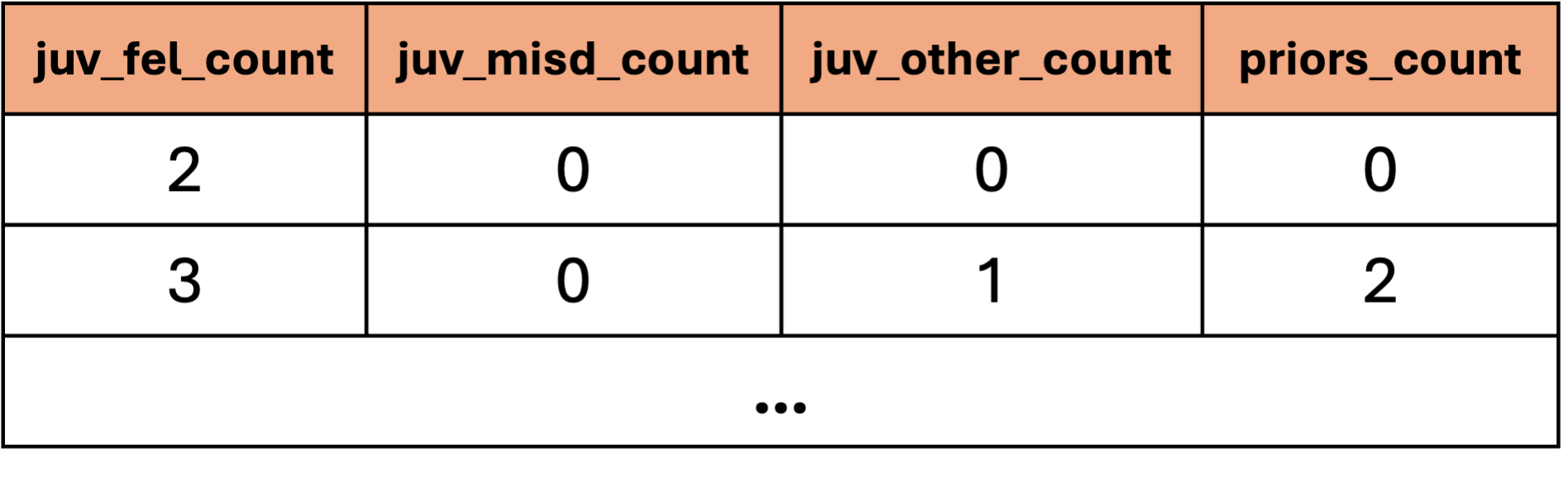}
        \caption{Single LLM.}
        \label{output_comp_llm_single}
    \end{subfigure}%
    \begin{subfigure}{0.49\textwidth} 
        \centering
        \includegraphics[width=\linewidth]{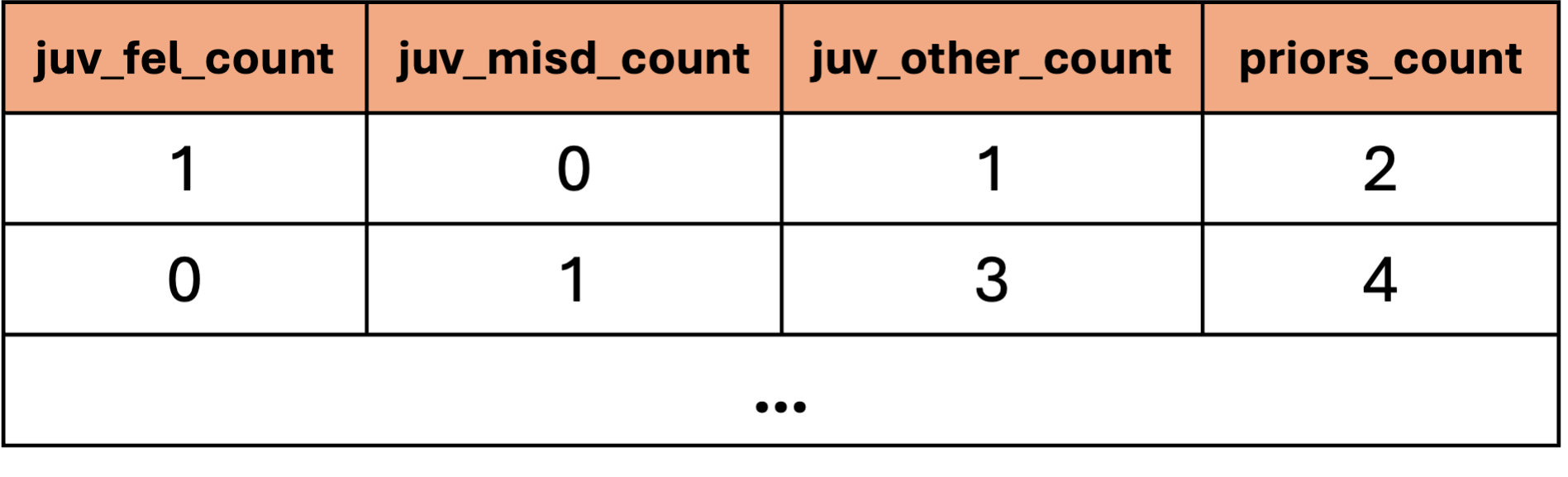}        
        \caption{LLM teaming.}
        \label{output_comp_llm_team}
    \end{subfigure}%
    \caption{Comparison of generated data on COMPAS from a single LLM and the proposed LLM teaming framework.}
    \label{output_comp_llm}
\end{figure}

\section{Example: The Advantage of LLM Teaming}\label{expteam}

Figure \ref{output_comp_llm} presents generated data examples from a single LLM and from LLM teaming on COMPAS dataset. The feature ``juv\_fel\_count" denotes the number of juvenile felonies of a defendant, while ``priors\_count" represents the total number of prior criminal records. By definition, ``priors\_count" should always be greater than or equal to ``juv\_fel\_count". As shown in Figure \ref{output_comp_llm_single}, data generated by a single LLM fails to respect this rule, whereas in Figure \ref{output_comp_llm_team}, our proposed LLM teaming framework successfully produces data that adheres to such basic consistency constraints. Our LLM teaming decomposes the task into semantically coherent subtasks, which reduces the prompt complexity and allows the LLM to generate more consistent and domain-aligned data. Worker LLMs condition on previously generated components to preserve inter-component logic, which is essential when domain constraints span multiple columns. Although we use the same LLM backbone, the role-specific instructions enforce separation of tasks, e.g., the label generator only reasons about outcome distribution, while other workers focus strictly on semantically related subsets of predictors. This is difficult to achieve with a single LLM, even with sophisticated prompting.


\end{document}